\documentclass[journal]{IEEEtran}
\usepackage{cite}
\usepackage{amsmath,amssymb,amsfonts}
\usepackage{algorithmic}
\usepackage{graphicx}
\usepackage{textcomp}
\usepackage{amsmath,amsfonts}
\usepackage{algorithmic}
\usepackage{algorithm}
\usepackage{array}
\usepackage[caption=false,font=normalsize,labelfont=sf,textfont=sf]{subfig}
\usepackage{textcomp}
\usepackage{stfloats}
\usepackage{url}
\usepackage{verbatim}
\usepackage{graphicx}
\usepackage{cite}
\usepackage{graphicx}
\usepackage{amsmath}
\usepackage{amssymb}
\usepackage{booktabs}
\usepackage{array,multirow,graphicx}
\usepackage{pifont}
\usepackage{setspace}
\usepackage{bbm}
\usepackage{caption}
\usepackage{comment}
\usepackage{tabularx}
\usepackage{booktabs}

\usepackage{hyperref}
\usepackage[dvipsnames]{xcolor}

\usepackage{xcolor}
\usepackage[normalem]{ulem}

\newcommand{\FH}[1]{\textcolor{black}{#1}}

\ifCLASSINFOpdf
\else
\fi

\begin{document}

\title{\FH{Few-shot Object Detection in Remote Sensing: Lifting the Curse of Incompletely Annotated Novel Objects}}

\author{Fahong Zhang, Yilei Shi,~\IEEEmembership{Member,~IEEE,}
Zhitong Xiong,~\IEEEmembership{Member,~IEEE,}
and Xiao Xiang Zhu,~\IEEEmembership{Fellow,~IEEE,}
\thanks{The work is jointly supported by the German Research Foundation (DFG GZ: ZH 498/18-1; Project number: 519016653), the Helmholtz Association through the Framework of the Helmholtz Excellent Professorship ``Data Science in Earth Observation - Big Data Fusion for Urban Research''(grant number: W2-W3-100), by the German Federal Ministry of Education and Research (BMBF) in the framework of the international future AI lab "AI4EO -- Artificial Intelligence for Earth Observation: Reasoning, Uncertainties, Ethics and Beyond" (grant number: 01DD20001) and by German Federal Ministry for Economic Affairs and Climate Action in the framework of the "national center of excellence ML4Earth" (grant number: 50EE2201C).}
\thanks{(Correspondence: Xiao Xiang Zhu)}
\thanks{
F. Zhang, Z. Xiong and X. Zhu are with the Chair of Data Science in Earth Observation, Technical University of Munich, 80333 Munich, Germany (e-mail: (fahong.zhang, zhitong.xiong, xiaoxiang.zhu)@tum.de).
}
\thanks{Y. Shi is with the Chair of Remote Sensing Technology, Technical University of Munich (TUM), 80333 Munich, Germany
(e-mail: yilei.shi@tum.de)}}

\markboth{IEEE Transactions on Geoscience and Remote Sensing ,~Vol.~13, No.~9, December~2023}%
{Shell \MakeLowercase{\textit{et al.}}: Bare Demo of IEEEtran.cls for Journals}

\maketitle

\begin{abstract}
Object detection is an essential and fundamental task in computer vision and satellite image processing. Existing deep learning methods have achieved impressive performance thanks to the availability of large-scale annotated datasets. Yet, in real-world applications the availability of labels is limited. In this context, few-shot object detection (FSOD) has emerged as a promising direction, which aims at enabling the model to detect novel objects with only few of them annotated. However, many existing FSOD algorithms overlook a critical issue:
when an input image contains multiple novel objects and only a subset of them are annotated,
the unlabeled objects will be considered as background during training.
This can cause confusions and severely impact the model's ability to recall novel objects. 
To address this issue, we propose a self-training-based FSOD (ST-FSOD) approach, which incorporates the self-training mechanism into the few-shot fine-tuning process. ST-FSOD aims to enable the discovery of novel objects that are not annotated, and take them into account during training. 
On the one hand, we devise a two-branch region proposal networks (RPN) to separate the proposal extraction of base and novel objects, On another hand, we incorporate the student-teacher mechanism into \FH{RPN and the region of interest (RoI) head} to include those highly confident yet unlabeled targets as pseudo labels. 
Experimental results demonstrate that our proposed method outperforms the state-of-the-art in various FSOD settings by a large margin. The codes will be publicly available at \url{https://github.com/zhu-xlab/ST-FSOD}.
\end{abstract}


\section{Introduction}
\label{sec:introduction}
\begin{figure*}[htp]
    \centering
    \includegraphics[width=0.8\linewidth]{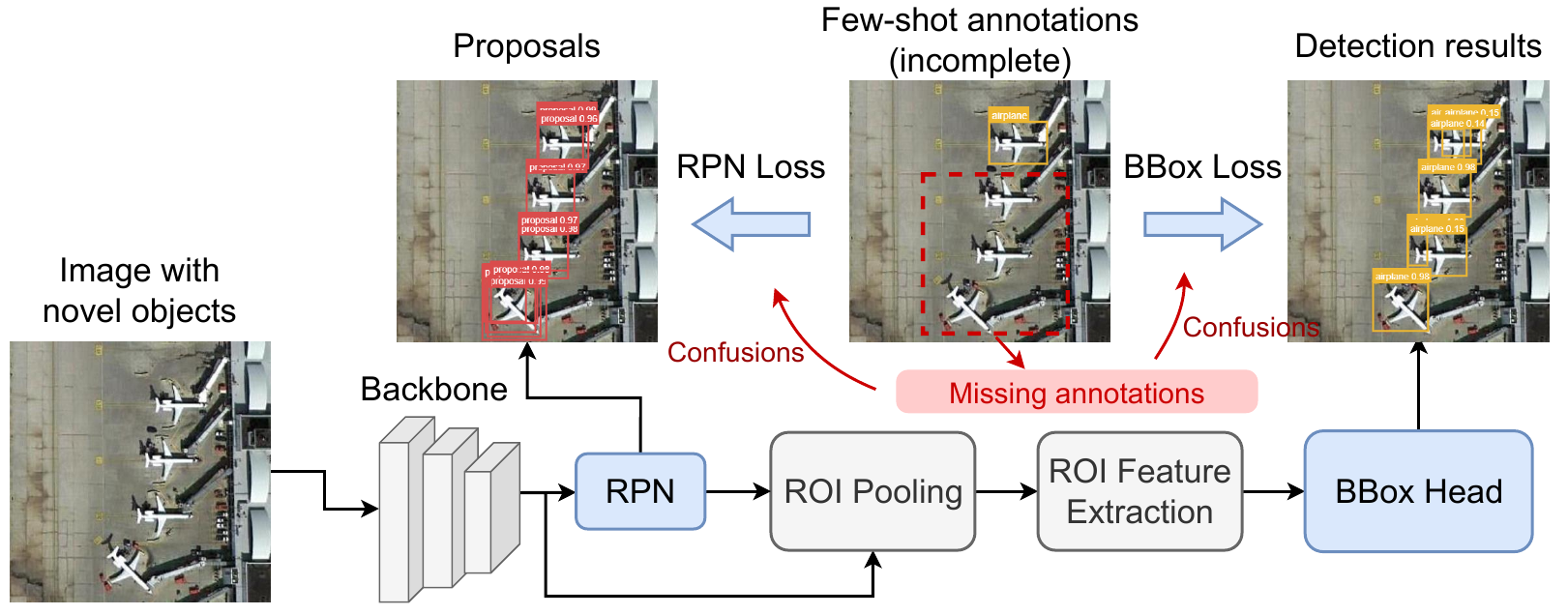}
    \caption{
        \FH{Illustration of the incompletely annotated novel objects (IANO) issue.
        In the standard FSOD protocal, only a few bounding boxes for the novel class object provided for the few-shot fine-tuning stage \cite{huang2022survey}.
        Let's consider a scenario where our goal is detecting the presence of a "plane" as the novel class object, and we are provided with just one bounding box annotation for a plane.
        As depicted in the figure,
        the challenge arises when multiple instances of planes are present within a single image,
        In such cases, some of the planes within the image are left unannotated.
        This can mislead the detector since RPN loss and bounding box classification loss are calculated based on these incomplete annotations.
        }
    }
    \label{fig:motivation}
\end{figure*}

Object detection (OD) is a critical task in computer vision as well as remote sensing image processing, which enables the automatic identification and localization of objects of interest within an image.
With the rapid development of deep learning techniques \cite{ren2015faster,redmon2016you,carion2020end} and the emergence of large-scale human-annotated data \cite{5206848,lin2014microsoft}, the performance of the state-of-the-art OD approaches has been pushed to a new stage. These approaches have achieved remarkable success in detecting objects in various domains, including remote sensing \cite{10075555,dai2022ao2,zhang2023semi}. However, traditional OD methods rely on a large amount of labeled data for training, which can be challenging and time-consuming to obtain in remote sensing image processing, particularly in scenarios where novel or rare objects are involved. 

\FH{
This results in the need for few-shot learning \cite{ wang2020generalizing,lang2022learning, xiong2022doubly}, a paradigm that aims to overcome the data scarcity issue by enabling models to generalize and detect objects with only a limited number of labeled examples.
Few-shot learning achieves this by leveraging knowledge acquired from previously seen categories to adapt and recognize novel objects efficiently.
}

\FH{Concurrently with the achievements in few-shot classification \cite{vinyals2016matching} and few-shot semantic segmentation \cite{cheng2022holistic,lang2023base}, few-shot object detection (FSOD) \cite{chen2018lstd,wang2020frustratingly,9884721} has emerged as a compelling research area in recent years. 
In the conventional FSOD framework, the model undergoes a two-stage training process: first, it is trained on a large-scale labeled dataset consisting of base objects, and subsequently, it is fine-tuned on a fine-tuning set with only a few labeled novel object instances. }

\FH{However, 
when there are multiple novel objects in a single image, it is possible that only a part of them are provided with labels during the fine-tuning stage.
As a result, these incomplete annotations can negatively impact the training towards novel classes and hinder the discovery of novel objects.
This issue, illustrated in Fig. \ref{fig:motivation}, can be referred as the incompletely annotated novel objects (IANO) issue.}

\FH{
While the challenge of IANO has been investigated in the field of FSOD for natural images \cite{qiao2023few,li2021few},
it still remains unexplored in remote sensing.
However, objects in the remote sensing images are usually smaller,
and scenes often exhibit higher levels of congestion,
particularly in contexts with vehicles, planes, ships, and so on.
This phenomenon can be readily observed from different OD datasets tailored for remote sensing applications \cite{waqas2019isaid}.
As a result, 
the IANO problem takes on an even more formidable and pressing nature when considered in the context of remote sensing.
}

\FH{
To tackle this issue, it is necessary to establish a mechanism capable of identifying and subsequently excluding potential unannotated novel objects during the background sampling process.
A promising solution that emerges for mitigating this concern is self-training,
a well-established technique in the field of domain adaptation \cite{farahani2021brief,wurm2019semantic,qingsong2023uni} and semi-supervised learning \cite{wei2021crest,zhang2023semi}.} 
This type of methods first generate pseudo labels on unlabeled data using a pre-trained model and then fine-tune the model using these pseudo labels.
\FH{The underlying philosophy of \emph{generating pseudo labels and leveraging the unlabeled data} aligns with our objective of identifying potential unannotated novel objects,
as they are hidden in the background and remain unannotated.
}
Therefore, we propose using self-training as a feasible approach to tackle this issue in remote sensing imagery.

We build our self-training-based FSOD (ST-FSOD) method based on a popular object detection framework: Faster RCNN \cite{ren2015faster}, which consists of two stages.
In the first stage,
a region proposal networks (RPN) is used to generate a set of candidate object proposals, each with a corresponding objectness score and a bounding box regression score.
In the second stage,
the candidate object proposals are refined and classified using a fully-connected layer network, also known as a Region of Interest (RoI) pooling layer \cite{girshick2015fast}.

We incorporate the self-training mechanism into the RPN and the bounding box head (BBH) of the RoI layer,
and devise a self-training RPN (ST-RPN) and a self-training BBH (ST-BBH) modules accordingly.
In these two modules,
a momentum-based teacher-student modeling strategy \cite{araslanov2021self} is used to filter out highly confident potential novel class objects and refine the loss calculation.
More specifically, the ST-RPN module uses the teacher-student modeling strategy to filter out highly confident proposals for potential novel classes. Meanwhile, the ST-BBH module filters out highly confident novel bounding boxes. Both modules refine the loss calculation to improve the accuracy of the detection model.
Our contributions can be summarized as follows:
\begin{itemize}
    \item Our study highlights and examines the challenge of the \FH{IANO} issue in FSOD for remote sensing imagery. 
    To the best of our knowledge, this is the first work that discusses and tackles the issue in the field of remote sensing, which is neglected in many existing FSOD methods for remote sensing imagery and needs to be addressed to advance the field. 
    \item To handle the \FH{IANO} issue in object detection, we propose to apply self-training technique.
    To this end, ST-RPN and ST-BBH modules are devised to identify proposals or bounding boxes that are likely to include a novel object, even in the absence of novel class annotations.
    \item We conduct extensive experiments on \FH{three publicly available datasets}, and evaluate our proposed method under various FSOD settings. Experimental results demonstrate that our approach outperforms the state-of-the-art methods by a significant margin.
\end{itemize}

\section{Related Works}

\begin{figure*}[htp]
    \centering
    \includegraphics[width=0.92\linewidth]{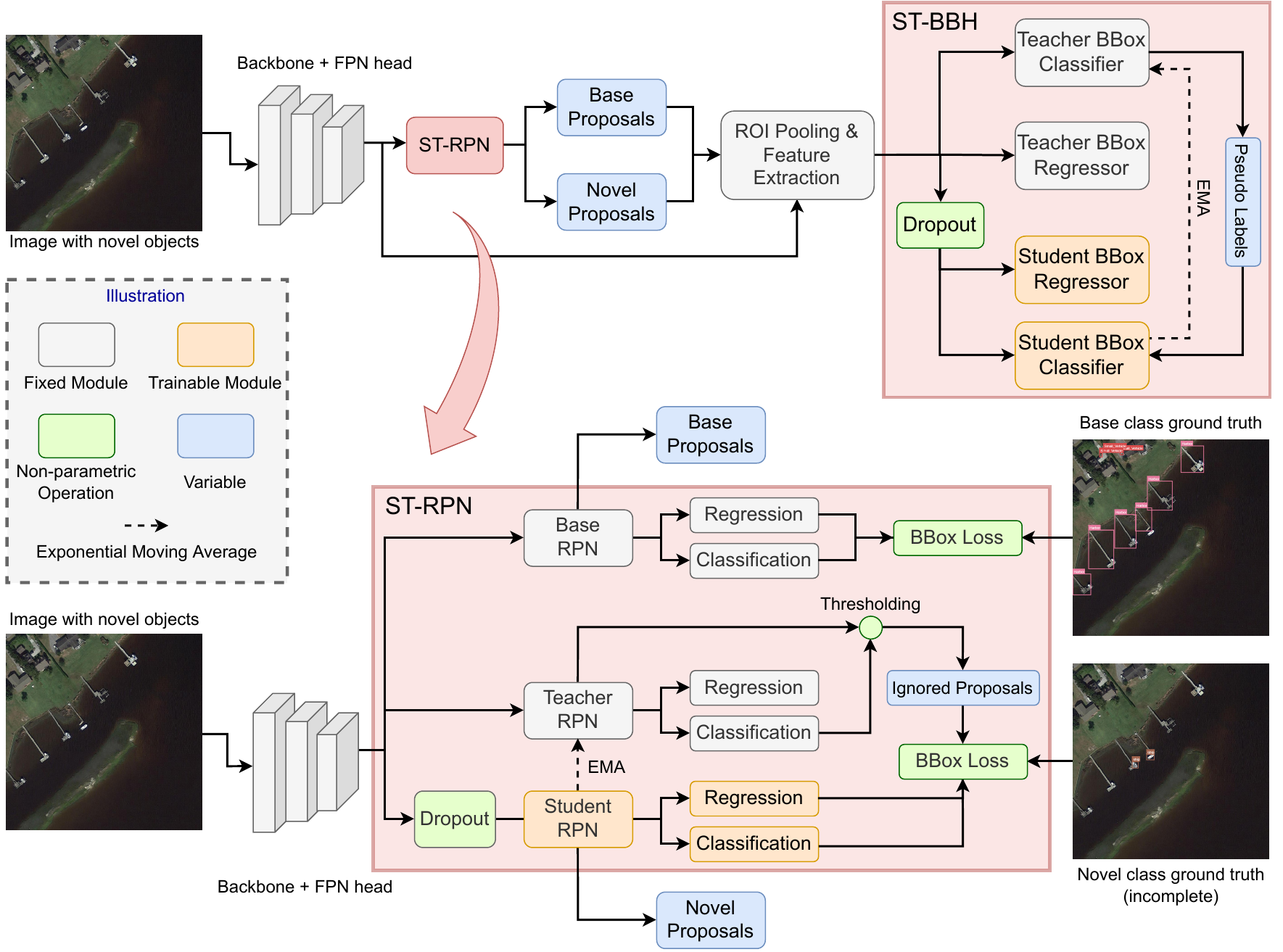}
    \caption{
        The overall architecture of the proposed method.
        The FPN following the backbone is not illustrated in the figures for the sake of simplicity.
    }
    \label{fig:architecture}
\end{figure*}

\subsection{Object Detection}
Object detection refers to identifying and localizing objects within an image, which has been one of the main research tasks in computer vision (CV). Traditional OD techniques are based on, e.g. feature extraction \cite{viola2001rapid}, object recognition, and template matching \cite{thanh2010improved}. State-of-the-art OD techniques are mostly based on deep learning, owing to its overwhelming performance on large-scale object detection benchmarks. Deep learning-based OD methods utilize convolutional neural networks (CNNs) to perform object detection directly from raw image pixels, without the need for hand-crafted feature engineering. They can be categorized into two-stage and single-stage detectors. While two-stage detectors aim to first generate object proposals and then classify them, single-stage detectors perform both tasks simultaneously. A typical example of two-stage detectors is Faster R-CNN \cite{ren2015faster}, which generates a set of region proposals and then extracts features from each proposal using a Region of Interest (RoI) pooling layer before classifying the object within the proposal.  One well-known single-stage detector is YOLO \cite{redmon2016you}, which applies a CNN to the entire image to simultaneously predict bounding boxes and class probabilities for each object without any proposal generation step.


\subsection{Few-shot Object Detection in Computer Vision}
Few-shot object detection (FSOD) aims at recognizing novel or unseen object classes based only on a few examples of them, by fine-tuning on a model trained on many labeled examples of base classes.
FSOD methods can be roughly categorized into fine-tuning-based methods, meta-learning-based methods, and metric-learning-based methods.

\subsubsection{Fine-tuning-based Methods}
Fine-tuning-based methods are popular for few-shot object detection, which first train on a large number of base class examples and then perform few-shot fine-tuning on a smaller support set that includes both base and novel classes.

One such method, LSTD \cite{chen2018lstd}, uses a flexible deep architecture that integrates the advantages of both SSD and Faster R-CNN in a unified deep framework. It also introduces transfer knowledge (TK) and background depression (BD) regularizations to leverage object knowledge from source and target domains during the fine-tuning stage.

Another fine-tuning-based method, TFA \cite{wang2020frustratingly}, indicates that even a simple fine-tuning of only the last layer of a Faster R-CNN detector to novel classes can achieve better performance compared to previous meta-learning based methods. In addition, TFA replaces the fully-connected classification heads of Faster R-CNN with cosine similarities. The aim is to  reduce intra-class variances and preserve performance in base classes through this feature normalization technique.

In \cite{yang2022efficient}, the authors introduce a pretrain-transfer framework (PTF) that utilizes a knowledge inheritance approach to initialize the weights for the box classifier. Additionally, they develop an adaptive length re-scaling strategy to ensure consistency of the dimensions of the pretrained weights for both the base and novel classes. This helps to improve the efficiency and effectiveness of the fine-tuning process.

\subsubsection{Metric-learning based Methods}

Metric-learning based methods aim to reduce the dimensionality of each sample and learn a feature representation such that similar samples are closer to each other, while dissimilar ones are easier to discriminate.

RepMet \cite{karlinsky2019repmet} is a metric-learning approach suitable for both few-shot classification and object detection. It uses a collection of Gaussian mixture models, each with multiple modes, to describe the base and novel classes. During base training, an embedding loss is employed to ensure a margin between the distance of each query feature and its respective class representative, as well as the distance to the nearest representative of an incorrect class.

In \cite{fan2020few}, the authors introduce an attention-based Region Proposal Network (attention-RPN) that uses support features to enhance the proposal generation process and eliminate non-matching or background proposals. Additionally, the authors develop a multi-relation detector for feature representation learning that measures the similarity between the RoI features of query and support objects.

\subsubsection{Meta-learning based Methods}

Meta-learning based methods are designed to learn quickly from a few examples, using a meta-learner that has been trained on a diverse set of tasks. During the base training stage, the meta-learner is trained on a meta-dataset composed of various tasks. Once trained, the meta-learner can quickly adapt to new tasks or generate a learner that is customized to the target task.

One example of meta-learning based methods is Meta-YOLO \cite{kang2019few}, which improves the query feature of a model by using a set of weighting coefficients generated during the meta-learning phase. These coefficients are based on the support samples and allow the model to effectively learn the intrinsic importance of features for detecting objects. This enables the re-weighting coefficients of novel classes to be learned with only a few support examples.

Another approach is FsDetView \cite{xiao2022few}, where a novel technique for aggregating query and support features is introduced. Instead of feature re-weighting, this technique involves performing element-wise multiplication, subtraction, and concatenation between the two sets of features. This approach has shown promising results in few-shot object detection tasks.

\subsection{Few-shot Object Detection in Remote Sensing}

Remote sensing images have unique characteristics that distinguish them from natural scene images, such as complex backgrounds, objects with multiple orientations, and dense and small objects. Thus, designing a few-shot object detection (FSOD) algorithm that accounts for these distinct features is crucial.

In SAM\&BFS \cite{huang2021few}, the authors propose a shared attention module that leverages class-agnostic prior knowledge gained during the base training stage to aid in detecting novel objects with significant size variations. In DH-FSDet \cite{wolf2021double}, the authors suggest using a balanced sampling strategy between base and novel samples, taking all the base samples into consideration. Additionally, they propose separating the classification and regression heads in the Region of Interest (RoI) layer according to base and novel classes for better balance in detection.
\FH{Zhang et al. \cite{zhang2023generalized} introduce the generalized FSOD task to remote sensing.
In one hand, they propose a metric-based discriminative loss to reduce the intra-class diversity and increase the inter-class separability of the base and novel objects.
In the other hand, they replace the RoI feature extractor with a representation compensation module to prevent the model from catastrophic forgetting.}

More recently, researchers have explored integrating text data into the visual learning pipeline to improve FSOD performance. Models such as TEMO \cite{lu2023few} and TSF-RGR \cite{zhang2023text} leverage text-modal knowledge extractors to provide prior knowledge on the relationship between base and novel classes, resulting in improved FSOD performance.

\subsection{Self-training}

Self-training \cite{yang2020fda,subhani2020learning,mei2020instance,shin2020two} is a widely used approach for domain adaptation \cite{10141556} and semi-supervised learning \cite{wei2021crest}. This technique involves generating pseudo-labels for the target domain or unlabeled data and then using them to fine-tune the network.
Self-training has shown promising results in various tasks, including semi-supervised classification \cite{berthelot2019mixmatch} and semantic segmentation \cite{hoyer2022daformer}.

In semi-supervised classification, FixMatch \cite{sohn2020fixmatch} is a popular approach for self-training. It generates pseudo-labels by using a model's predictions on weakly-augmented, unlabeled images. These pseudo-labels are then used to supervise the model's predictions on a strongly-augmented view of the same image. This approach has been shown to achieve state-of-the-art performance on several benchmark datasets.

For semantic segmentation tasks, SAC \cite{araslanov2021self} is a self-training approach that employs a momentum network as the teacher network. The teacher network generates pseudo-labels on the weakly-augmented images, which are used to supervise the student network that receives the strongly-augmented images. The momentum network is updated based on the exponential moving average of the student network. This approach has also shown promising results on several benchmark datasets.

\FH{
\subsection{Incompletely Annotated Novel Objects Issue}
Li et al. \cite{li2021few_2} are the first to identify the Icompletely Annotated Novel Objects (IANO) issue in their work, pointing out that the base set images might contain unlabeled novel-class objects, leading to false positives. They address this challenge by introducing a distractor utilization loss.
More specifically, for each annotated bounding box, a cropped image centered at the object will be fed into a few-shot correction network. This network generate corresponding pseudo labels, which are subsequently used to identify potential novel objects and adjust the loss calculation with respect tothe Region of Interest (RoI) head.
On a similar note, Qiao et al. \cite{qiao2023few} emphasize that the IANO issue also exists when multiple novel objects were present within a single image. To resolve this concern, they propose a label calibration method. This method recalibrated the predicted targets of background objects based on their predicted confidences. As a result, unannotated novel objects are assigned with lower weights during the loss calculation, mitigating their negative impact.}

\FH{
Despite these valuable contributions, it's worth noting that the aforementioned approaches primarily addressed the impact of unannotated objects to bounding box classification, neglecting their influence on the training of region proposal networks (RPN) for a two-stage object detector.
Specifically, when calculating the RPN loss, the incompleteness of novel object annotations could cause the RPN to mistakenly predict unannotated novel objects as background, thereby reducing overall performance.
In our method, we propose a comprehensive approach to mitigate the IANO issue. We apply advanced self-training techniques not only to the bounding box classification head but also to the RPN.
By doing so, we extend our efforts to tackle this challenge on a broader scale, addressing the impact of unannotated objects throughout the entire object detection pipeline.
}

\section{Methods}

In this section, we introduce the proposed ST-FSOD (Self-Training-based Few-Shot Object Detection) method, which consists of two major components: the Self-Training Region Proposal Network (ST-RPN) and the Self-Training Bounding Box Head (ST-BBH). The overall architecture is illustrated in Figure \ref{fig:architecture}.
The ST-RPN module takes the multi-level features extracted from the backbone and feature pyramid networks (FPN) head \cite{lin2017feature} as input and generates two sets of object proposals, namely, the base and novel proposals, corresponding to the base and novel categories, respectively. These proposals are merged and fed into the ROI pooling and feature extraction layers to obtain the ROI features.
The ST-BBH module takes the ROI features as input and produces the final detection results. Specifically, it detects potential \FH{unannotated novel class objects} and uses them as pseudo labels to recall more novel class objects, thereby improving the model's performance.

\subsection{Problem Formulation}
In this section, we present the formulation of the standard FSOD setting. We assume that we have access to a base set containing base class annotations denoted as $\mathcal{D}_{base}=\{(\mathbf{I}_i, \mathcal{B}_i^{base})\}$ and a novel set containing novel class annotations denoted as $\mathcal{D}_{nov}=\{(\mathbf{I}_i, \mathcal{B}_i^{nov})\}$, where $\mathbf{I}_i$ represents an image and $\mathcal{B}_i=\{(x,y,h,w,c)\}$ represents a set of object bounding boxes within the image. Here, $x,y,w$ and $h$ denote the locations of the bounding box, and $c$ denotes the class label.

Furthermore, let $\mathcal{C}_{base}$ and $\mathcal{C}_{nov}$ be the label set of $\mathcal{B}_{base}$ and $\mathcal{B}_{nov}$, respectively, and they satisfy the condition $\mathcal{C}_{base} \cap \mathcal{C}_{nov} = \varnothing$, indicating that there are no common classes between the base and novel classes.

Our proposed ST-FSOD method is established on the classic two-stage fine-tuning approach (TFA) \cite{wang2020frustratingly}.
In the first stage of TFA, a base detector is trained using the base set $\mathcal{D}_{base}$, following the same procedure as in a regular object detector. In the second stage, a few-shot object detector is initialized using the weights obtained in the first stage and fine-tuned on a $K$-shot fine-tuning set denoted by $\mathcal{D}_{ft} = {(\mathbf{I}_i, \mathcal{B}_i^{base}, \mathcal{B}_i^{nov})}$, where the number of novel object bounding boxes for each novel class is $K$. More formally, for each novel class $c_{nov} \in \mathcal{C}_{nov}$, we have:

\begin{equation}
\begin{aligned}
    &\forall c_{nov} \in \mathcal{C}_{nov}, \\
    &\sum_{\mathbf{I}_i, \mathcal{B}_i^{base}, \mathcal{B}_i^{nov} \in \mathcal{D}_{ft}} |\{(x, y, h, w, c) \in \mathcal{B}_i^{nov} | \ c = c_{nov}\}| = K.
\end{aligned}
\end{equation}
Here $|\cdot |$ denotes the number of elements of the set.

\subsection{Balanced Sampling Strategy}
\label{sec:method_bss}
One of the key questions in constructing the few-shot fine-tuning set $\mathcal{D}_{ft}$ is how many base class objects to sample.
In the original TFA \cite{wang2020frustratingly}, the authors proposed sampling exactly $K$ shots of base objects for each base class to maintain a better balance between base and novel classes.
However, recent studies have shown that this strategy may not be the optimal for remote sensing images \cite{wolf2021double,huang2021few}.
For example, Wolf et al. \cite{wolf2021double} found that using more base objects and oversampling the few-shot novel objects can improve overall performance.

Inspired by these findings, we propose a balanced sampling strategy (BSS) as follows:
First, we randomly sample the $K$-shot novel objects as usual.
Next, we include all base class images and annotations in $\mathcal{D}_{base}$ into the fine-tuning set $\mathcal{D}_{ft}$.
Finally, when sampling images for fine-tuning, we increase the probability of sampling images $\mathbf{I}_{i}$ with a corresponding non-empty $\mathcal{B}_i^{nov}$ to ensure they are sampled with the same probability as images that do not contain any novel annotations.
By adopting the BSS, we achieve a more balanced fine-tuning set between base and novel classes, while also making full use of all the available base annotations.

\subsection{Self-training-based Region Proposal Networks}
\label{sec:method_st_rpn}

In the Faster R-CNN network architecture, the Region Proposal Network (RPN) is utilized to generate a set of proposals $\mathcal{P}=\{p=(t_x, t_y, t_w, t_h, o)\}$ for each input image $I$ using multi-scale features extracted from $I$. The parameters $t_x$, $t_y$, $t_w$, and $t_h$ represent the coordinates of each proposal, and $o$ is an objectness score that indicates the probability of the proposal containing an object. Fig. \ref{fig:architecture} illustrates the architecture of ST-RPN, which consists of three sub-modules: base RPN, teacher RPN, and student RPN. All of these sub-modules follow the original RPN architecture. 
ST-RPN generates two sets of proposals, $\mathcal{P}^{base}$ and $\mathcal{P}^{nov}$, which are obtained as follows:

\begin{itemize}
    \item The base RPN is responsible for extracting base proposals $\mathcal{P}^{base}$ from the input image $\mathbf{I}$.
    For fine-tuning the base RPN module, only base annotations $\mathcal{B}^{base}$ are used to calculate the RPN loss.
    \begin{equation}
        \mathcal{L}_{rpn}^{base}(\mathcal{P}^{base}, \mathcal{B}^{base}) =
        \sum_{p_i \in \mathcal{P}^{base}} L(p_i, p_{i}^{*}),
    \end{equation}
    where $p_i^*$ is the regression and classification targets achieved by matching $\mathcal{P}^{base}$ with $\mathcal{B}^{base}$ according to the intersection over union (IoU) between each pair of the proposal and the ground truth bounding box \cite{ren2015faster}.

    \item The teacher RPN generates a set of ignored proposals $\mathcal{P}^{ign}$, which includes output proposals from this module that have an objectness score $o$ greater than a given threshold $\tau_{rpn}$.
    Please refer to \ref{sec:expr_details} for the selection of $\tau_{\text{rpn}}$.

    \item The student RPN receives the extracted features from the backbone and the FPN head \cite{lin2017feature} with dropout \cite{srivastava2014dropout} applied and outputs a set of novel proposals $\mathcal{P}^{nov}$. This module is trainable and is supervised only by the few-shot novel annotations $\mathcal{B}^{nov}$. During the calculation of the regression and classification target of each output proposal, those with a high overlap with the ignored proposals in $\mathcal{B}^{ign}$ are excluded from the loss calculation.
    The loss function for the student RPN can be formulated as follows:
    \begin{equation}
        \mathcal{L}_{rpn}^{nov}(\mathcal{P}^{nov}, \mathcal{P}^{ign}, \mathcal{B}^{nov}) =
        \sum_{p_i \in \mathcal{P}^{nov}} w_i L(p_i, p_{i}^{*}).
    \end{equation}
    Here $p_i^*$ is the box targets achieved by assigning $\mathcal{P}^{nov}$ to $\mathcal{B}^{nov}$.
    $w_i$ is a weighting coefficient that is used to ignore the highly confident proposals that might contain novel objects.
    \begin{equation}
        \begin{aligned}
            w_i = &  \begin{cases}
            0, & \text{if } \exists \ p_j \in \mathcal{P}^{ign}, \text{IoU}(p_j, p_i) > 0.7, \\
            1, & \text{otherwise}.
        \end{cases}
        \end{aligned}
    \end{equation}
    Here $\text{IoU}(\cdot, \cdot)$ denotes the IoU value of the two proposals.
\end{itemize}

The separation of base and novel proposals has two benefits. First, during fine-tuning, the pre-trained weights obtained from the base training stage could be negatively affected, which can impact the quality of the extracted base proposals. Separating proposal extraction prevents the fine-tuning towards novel objects from biasing the extraction of base proposals. Second, previous FSOD works in remote sensing \cite{huang2021few} have found that the RPN module achieved from the base training stage often fails to recall novel objects successfully. Thus, training an additional RPN module from scratch can lead to a better network state in terms of novel object detection.

During the extraction of novel proposals $\mathcal{P}^{nov}$, using a student-teacher-based self-training mechanism and introducing ignored proposals into the loss calculation can prevent potential unannotated novel objects (UNO) from being misclassified as background. This helps the RPN module to recall more novel objects, and improves detection performance.

\subsection{Self-training-based Bounding Box Head}
In the proposed framework, the extracted base and novel proposals are merged and passed through a RoI pooling layer and a RoI feature extraction layer to obtain the corresponding RoI features. These features are then forwarded to the self-training bounding box head (ST-BBH) for achieving the final detection results.
The ST-BBH consists of a teacher and a student bounding box heads (BBH). Each BBH contains a bounding box classifier and a regressor, following the same architecture as the one proposed in Faster R-CNN \cite{girshick2015fast}.
While processing each RoI, the feature is input to both the teacher and student classifier heads. 
In order to improve the robustness, the student head receives the feature with dropout \cite{srivastava2014dropout} applied as the input.
Let $\mathbf{u}^{stu}$ and $\mathbf{u}^{tch}$ be the output probability of the student and teacher BBH, and $v$ be the corresponding ground truth label.
The student classifier's classification loss is calculated using the following equation:

\begin{equation}
\begin{aligned}
&\mathcal{L}_{bbh}(\mathbf{u}^{\text{stu}}, \mathbf{u}^{\text{tch}}, v) = \\
&\quad \ \begin{cases}
L_{\text{cls}}(\mathbf{u}^{\text{stu}}, \hat{u}^{\text{tch}}), & \text{if } v = 0 \text{ and } \max(\mathbf{u}^{\text{tch}}) > \tau_{\text{bbox}}, \\
L_{\text{cls}}(\mathbf{u}^{\text{stu}}, v), & \text{otherwise}.
\end{cases}
\end{aligned}
\end{equation}
Here, $\hat{u}^{\text{tch}}$ represents the class index with the highest value in $\mathbf{u}^{\text{tch}}$.
The assigned ground truth class label of the RoI is denoted by $v$, with $v=0$ indicating that the RoI is considered as background. The threshold $\tau_{\text{bbox}}$ is used to determine when to use the prediction from the teacher bounding box head as the pseudo label.
Please refer to \ref{sec:expr_details} for the selection of $\tau_{\text{bbox}}$.

\FH{
Overall, ST-RPN and ST-BBH share the same self-training philosophy with a momentum network, represented in our context as the teacher module. This teacher module maintains a slowly updated copy of the original module, ensuring stable yet recent targets (or pseudo-labels) for model updates, as discussed in \cite{araslanov2021self}.
However, it's important to note that ST-RPN and ST-BBH are technically distinct: ST-RPN is employed to extract class-agnostic proposals, while ST-BBH is specifically designed for the classification and regression of class-specific bounding boxes.
Furthermore, we emphasize that in ST-RPN, we explicitly separate the extraction of base and novel proposals, whereas in ST-BBH, the classification and regression heads for both base and novel classes share the same ROI features.
}

\subsection{Weights Initialization \& Update}

As depicted in Fig. \ref{fig:architecture}, different trainable and fixed modules have been highlighted in different colors.
Among these modules, the backbone, FPN head, base RPN, and RoI feature extraction layer will be initialized by the pre-trained weights obtained from the base training stage.
\FH{
For the classifier and regressor of both student and teacher BBH, the entries of their weight matrices correspond to the base classes will be initialized based on the pre-trained weights, while entries for the novel classes will be randomly initialized. The student and teacher RPN module of the ST-RPN will be randomly initialized.}

It is worth noting that the weights for the teacher RPN and teacher BBH are not trainable. Instead, they will be updated by the corresponding student module's weights using exponential moving average \cite{araslanov2021self}.
\begin{equation}
   \theta_{\mathcal{T}}^{(t)} = \alpha \theta_{\mathcal{T}}^{(t-1)} + (1 - \alpha) \theta_{\mathcal{S}}^{(t)},
\end{equation}
Here $\theta_{\mathcal{T}}^{(t)}$ and $\theta_{\mathcal{S}}^{(t)}$ denote the weights of the teacher networks and their corresponding student networks at time step $t$ during the training stage, respectively.
$\alpha$ is a decay weight that is set to $0.999$ following \cite{araslanov2021self}.

\section{Experiments}
In this section, we will present the experimental results of our proposed method on various benchmarks for FSOD in remote sensing.

\subsection{Experimental Settings}
\label{sec:expr_setting}

We evaluate the proposed method on \FH{three} large-scale public object detection datasets in remote sensing, including \FH{NWPU-VHR10 v2 \cite{cheng2014multi}}, DIOR \cite{li2020object} and iSAID \cite{waqas2019isaid}.
\FH{
NWPU-VHR10 v2 dataset comprises a total of $1,172$ images,
each with dimension $400 \times 400$, and are divided into $10$ categories.
Following the previous works \cite{zhang2021oriented},
$3$ categories ``airplane'', ``baseball dimond'', and ``tennis court'' are adopted as novel classes,
while the others are as the base classes.
In line with previous researches, we employ the training and validation sets to fine-tune the model, and report the performance on the test set, which contains $293$ images.
}

The DIOR dataset contains $23,463$ images and over $190,000$ instances, with an image size of $800 \times 800$.
All objects are categorized into 20 classes.
In previous literature, two commonly used settings are adopted.
The first setting, proposed by Li et al. \cite{li2021few}, uses $5$ categories (i.e., ``plane'', ``baseball field'', ``tennis court'', ``train station'', and ``wind mill'') as the novel categories and the remaining as the base categories.
In this setting, the training set is used for base training and few-shot fine-tuning, while the validation set is used for evaluation.
The second setting, proposed in \cite{cheng2021prototype}, includes a total of $4$ base-novel class splits, each containing $5$ novel categories and $15$ base categories.
In this setting, both the training and validation sets are used for base training and fine-tuning, and the test set is used for evaluation.

iSAID is a large-scale instance segmentation dataset for remote sensing.
It is built on the same image set as DOTA \cite{xia2018dota},
but provide the instance-level mask annotations,
and also finer bounding box annotations.
iSAID contains 2806 images, whose sizes range from $800 \times 800$ to $20,000 \times 20,000$.
In total, there are $655,451$ annotated objects,
which are classified into $15$ categories.
We follow the official data pre-processing pipeline to crop the images into $800 \times 800$ patches, with a overlap of $25\%$.
We follow the FSOD setting of \cite{wolf2021double},
which uses $3$ different base-novel class splits,
and sets the number of shots for each split to $10$, $50$, and $100$.

To make a fair comparison with the previous works,
we adopt the mean average precision (mAP) with a IoU threshold at $0.5$ as the evaluation metric following the common practices.

\subsection{Implementation Details}
\label{sec:expr_details}

The proposed method uses the Faster R-CNN architecture \cite{ren2015faster} with a ResNet101 \cite{7780459} backbone that is pre-trained on the ImageNet dataset. A Feature Pyramid Networks (FPN) \cite{lin2017feature} is used to generate multi-scale features. The Adamw optimizer \cite{ilya2019decoupled} with a weight decay of 0.01 and a learning rate of $1e-4$ is used to train the model on all settings. 
\FH{The base training stage of NWPU-VHR10 v2, DIOR and iSAID datasets lasts for $10,000$, $40,000$ and $80,000$ iterations, respectively. 
Learning rate decay with a factor of 0.1 is applied at $5,000$ and $8,000$ for NWPU-VHR10 v2 dataset, $24,000$ and $32,000$ iterations for DIOR dataset, and $40,000$ and $60,000$ for iSAID dataset.
The few-shot fine-tuning lasts for $2,000$ iterations for NWPU-VHR10 v2 dataset and $10,000$ iterations for the other settings.
}

For data pre-processing and augmentation, image patches are randomly cropped to sizes of \FH{400 $\times$ 400 for NWPU-VHR10 v2 dataset and 608 $\times$ 608 for the others}.
Multi-scale training with a range from 0.5 to 2.0, random flipping, and random rotation with degrees of 90, 180, and 270 are applied. The batch size is set to \FH{16 for base training and 8 for fine-tuning.}

The momentum parameter is set to $\alpha=0.999$ when updating the networks' weights by exponential moving average \cite{araslanov2021self}. The thresholds $\tau_{\text{rpn}}$ and $\tau_{\text{bbh}}$ used in ST-ROI and ST-BBH are set to 0.8. In Sec. \ref{sec:expr_parameter}, sensitivity analyses to these hyperparameters are provided. 
Our codes are based on PyTorch, EarthNets \cite{xiong2022earthnets}, and MMDetection \cite{mmdetection} platform.
For more information, please refer to our published codes.

\setlength\tabcolsep{0pt}
\begin{table*}
\footnotesize
\caption{
    \FH{Average Precision (AP) (in $\%$) at a IoU threshold of $0.5$ of different methods on NWPU-VHR v2 dataset, where the base-novel class split follows \cite{zhang2021oriented}.
    Averaged results and the standard deviations of $3$ different runs are reported for the proposed methods.}
}
\renewcommand\arraystretch{1.2}
\newcolumntype{C}{>{\centering \arraybackslash}m{1.5cm}}
\newcolumntype{D}{>{\centering \arraybackslash}m{1.5cm}}
\begin{center}
    \begin{tabular}{>{\centering \arraybackslash}m{5cm} | C C C C | C C C C}
\toprule
        \multirow{2}{*}{\parbox{0.1\linewidth}{\vspace{0.1cm}}
        Method / Shots }
         & \multicolumn{4}{c|}{Novel Classes} & \multicolumn{4}{c}{Base Classes} \\
 & 3 & 5 & 10 & 20 & 3 & 5 & 10 & 20  \\

\addlinespace[0.5ex]
\cline{1-9}
\addlinespace[0.5ex]
OFA \cite{zhang2021oriented}  & 43.2 & 60.4 & 66.7 & - & - & - & - & -\\
FSODM \cite{li2021few}  & 32 & 53 & 65 & - & - & - & - & - \\
SAM\&BFS \cite{huang2021few}  & 47.0 & 61.6 & 74.9 & - & - & - & - & -\\
PAMS-Det \cite{zhao2021few} & 37 & 55 & 66 & - & - & - & - & - \\
CIR-FSD \cite{wang2022context} & 54 & 64 & 70 & - & - & - & - & - \\
TFACSC \cite{li2022few} & 47 & \bf{67} & 72 & - & - & - & - & - \\
SAGS\&TFS \cite{zhang2022few} & 51 & 66 & 72 & - & - &  - & - & - \\
TSF-RGR \cite{zhang2023text} & 57 & 66 & \bf{77} & - & - &  - & - & - \\
G-FSOD \cite{zhang2023generalized} & 50.1 & 58.8 & 67.0 & 75.9 & \bf{90.0} & \bf{90.5} & 89.2 & \bf{90.6} \\
\addlinespace[0.5ex]
\cline{1-9}
\addlinespace[0.5ex]
Ours & \bf{60.7} $\pm$ 2.1 & \bf{67.2} $\pm$ 1.2 & \bf{77.2} $\pm$ 2.5 & \bf{83.3} $\pm$ 1.5 & 89.7 $\pm$ 0.4 & 89.1 $\pm$ 0.5 & \bf{89.3} $\pm$ 1.1 & 89.9 $\pm$ 0.5 \\
\bottomrule
\end{tabular}
\end{center}
\label{tab:nwpu}
\end{table*}

\setlength\tabcolsep{0pt}
\begin{table*}
\footnotesize
\caption{
    Average Precision (AP) (in $\%$) at a IoU threshold of $0.5$ of different methods on DIOR dataset, where the base-novel class split follows \cite{li2021few}.
    Averaged results and the standard deviations of $3$ different runs are reported for the proposed methods.
}
\renewcommand\arraystretch{1.2}
\newcolumntype{C}{>{\centering \arraybackslash}m{1.5cm}}
\newcolumntype{D}{>{\centering \arraybackslash}m{1.5cm}}
\begin{center}
    \begin{tabular}{>{\centering \arraybackslash}m{5cm} | C C C C | C C C C}
\toprule
        \multirow{2}{*}{\parbox{0.1\linewidth}{\vspace{0.1cm}}
        Method / Shots }
         & \multicolumn{4}{c|}{Novel Classes} & \multicolumn{4}{c}{Base Classes} \\
 & 3 & 5 & 10 & 20 & 3 & 5 & 10 & 20  \\

\addlinespace[0.5ex]
\cline{1-9}
\addlinespace[0.5ex]
OFA \cite{zhang2021oriented}  & 32.8 & 37.9 & 40.7 & - & - & - & - & -\\
FSODM \cite{li2021few}  & - & 25 & 32 & 36 & - & - & - & - \\
SAM\&BFS \cite{huang2021few}  & - & 38.3 & 47.3 & 50.9 & - & - & - & -\\
PAMS-Det \cite{zhao2021few} & 28 & 33 & 38 & - & - & - & - & - \\
CIR-FSD \cite{wang2022context} & - & 33 & 38 & 43 & - & - & - & - \\
TFACSC \cite{li2022few} &  38 & 42 & 47 & - & - & - & - & - \\
SAGS\&TFS \cite{zhang2022few} & - & 34 & 37 & 42 & - &  - & - & - \\
TSF-RGR \cite{zhang2023text} & - & 42 & 49 & 54 & - & - & - & -\\
\addlinespace[0.5ex]
\cline{1-9}
\addlinespace[0.5ex]
Ours & \bf{43.5} $\pm$ 5.5 & \bf{48.3} $\pm$ 1.3 & \bf{55.8} $\pm $ 1.5 & \bf{61.3} $\pm$ 2.0 & 75.0 $\pm$ 0.3 & 75.6 $\pm$ 0.5 & 74.4 $\pm$ 0.3 & 73.2 $\pm$ 2.1 \\
\bottomrule
\end{tabular}
\end{center}
\label{tab:dior_part1}
\end{table*}

\setlength\tabcolsep{0pt}
\begin{table*}
\footnotesize
\caption{
    Average Precision (AP) (in $\%$) at a IoU threshold of $0.5$ of different methods on DIOR dataset, where the 4 base-novel class splits follow \cite{cheng2021prototype}.
    Averaged results and the standard deviations of $3$ different runs are reported for the proposed methods.
}
\renewcommand\arraystretch{1.2}
\newcolumntype{C}{>{\centering \arraybackslash}m{1.6cm}}
\newcolumntype{D}{>{\centering \arraybackslash}m{1.5cm}}
\begin{center}
    \begin{tabular}{C | >{\centering \arraybackslash}m{3cm} |  C C C C | C C C C}
\toprule
        \multirow{2}{*}{\parbox{0.1\linewidth}{\vspace{0.1cm}}
        Shots}
         & \multirow{2}{*}{\parbox{0.1\linewidth}{\vspace{0.1cm}}
        Methods}
        & \multicolumn{4}{c|}{Novel Classes} & \multicolumn{4}{c}{Base Classes} \\
 & & split1 & split2 & split3 & split4 & split1 & split2 & split3 & split4 \\

\addlinespace[0.5ex]
\cline{1-10}
\addlinespace[0.5ex]
\multirow{4}{*}{}{3} 
& P-CNN \cite{cheng2021prototype} & 18.0 & 14.5 & 16.5 & 15.2 & 47.0 & 48.9 & 49.5 & 49.8 \\
& SAGS\&TFS \cite{zhang2022few} & 29.3 & 12.6 & \bf{20.9} & 17.5 & - & - & - & - \\
& G-FSOD \cite{zhang2023generalized} & 27.6 & 14.1 & 16.0 & 16.7 & 68.9 & 69.2 & 71.1 & 69.0 \\
& Ours & \bf{41.9} $\pm$ 0.6 & \bf{17.7} $\pm$ 2.0 & \bf{20.9} $\pm$ 0.4 & \bf{20.4} $\pm$ 3.6 & \bf{73.5} $\pm$ 0.5 & \bf{72.5} $\pm$ 0.5 & \bf{75.2} $\pm$ 0.4 & \bf{73.3} $\pm$ 0.6\\
\addlinespace[0.5ex]
\cline{1-10}
\addlinespace[0.5ex]
\multirow{3}{*}{}
{5}
& P-CNN \cite{cheng2021prototype} & 22.8 & 14.9 & 18.8 & 17.5 & 48.4 & 49.1 & 49.9 & 49.9 \\
& SAGS\&TFS \cite{zhang2022few} & 31.6 & 15.5 & 24.8 & 19.7 & - & - & - & - \\
& G-FSOD \cite{zhang2023generalized} & 30.5 & 15.8 & 23.3 & 21.0 & 69.5 & 69.3 & 70.2 & 68.0 \\
& Ours & \bf{45.7} $\pm$ 1.6 & \bf{20.7} $\pm$ 2.8 & \bf{26.0} $\pm$ 2.5 & \bf{25.2} $\pm$ 4.5 & \bf{73.3} $\pm$ 0.4 & \bf{72.7} $\pm$ 0.4 & \bf{75.6} $\pm$ 0.6 & \bf{73.5} $\pm$ 0.4  \\
\addlinespace[0.5ex]
\cline{1-10}
\addlinespace[0.5ex]
\multirow{3}{*}{}
{10} 
& P-CNN \cite{cheng2021prototype} & 27.6 & 18.9 & 23.3 & 18.9 & 50.9 & 52.5 & 52.1 & 51.7 \\
& SAGS\&TFS \cite{zhang2022few} & 31.6 & 15.5 & 24.8 & 19.7 & - & - & - & - \\
& G-FSOD \cite{zhang2023generalized} & 37.5 & 20.7 & 26.2 & 25.8 & 69.0 & 68.7 & 71.1 & 68.6 \\
& Ours & \bf{50.0} $\pm$ 1.5 & \bf{27.3} $\pm$ 1.1 & \bf{31.3} $\pm$ 0.3 & \bf{33.4} $\pm$ 1.1 & \bf{72.6} $\pm$ 0.3 & \bf{72.3} $\pm$ 0.5 & \bf{75.7} $\pm$ 0.4 & \bf{73.9} $\pm$ 0.2 \\
\addlinespace[0.5ex]
\cline{1-10}
\addlinespace[0.5ex]
\multirow{3}{*}{}
{20} 
& P-CNN \cite{cheng2021prototype} & 29.6 & 22.8 & 28.8 & 25.7 & 52.2 & 51.6 & 53.1 & 52.3 \\
& SAGS\&TFS \cite{zhang2022few} & 40.2 & 23.8 & \bf{36.1} & 27.7 & - & - & - & - \\
& G-FSOD \cite{zhang2023generalized} & 39.8 & 22.7 & 32.1 & 31.8 & 69.8 & 68.2 & 71.3 & 67.7 \\
& Ours & \bf{53.7} $\pm$ 1.1 & \bf{33.4} $\pm$ 0.4 & 34.6 $\pm$ 1.9 & \bf{38.2} $\pm$ 2.0 & \bf{73.3} $\pm$ 0.5 & \bf{73.3} $\pm$ 0.5 & \bf{75.5} $\pm$ 0.2 & \bf{73.8} $\pm$ 0.2 \\
\bottomrule
\end{tabular}
\end{center}

\label{tab:dior_part2}
\end{table*}

\setlength\tabcolsep{0pt}
\begin{table*}
\footnotesize
\caption{
    Average Precision (AP) (in $\%$) at a IoU threshold of $0.5$ of different methods on iSAID dataset, where the 3 base-novel class splits follows \cite{wolf2021double}.
    Averaged results and the standard deviations of $3$ different runs are reported for the proposed methods. Results of FSDetView and TFA are cited from \cite{wolf2021double}.
}
\renewcommand\arraystretch{1.2}
\newcolumntype{C}{>{\centering \arraybackslash}m{1.6cm}}
\newcolumntype{D}{>{\centering \arraybackslash}m{1.5cm}}
\begin{center}
    \begin{tabular}{C | >{\centering \arraybackslash}m{3cm} |  C C C | C C C}
\toprule
        \multirow{2}{*}{\parbox{0.1\linewidth}{\vspace{0.1cm}}
        Shots}
         & \multirow{2}{*}{\parbox{0.1\linewidth}{\vspace{0.1cm}}
        Methods}
        & \multicolumn{3}{c|}{Novel Classes} & \multicolumn{3}{c}{Base Classes} \\
 & & split1 & split2 & split3 & split1 & split2 & split3 \\

\addlinespace[0.5ex]
\cline{1-8}
\addlinespace[0.5ex]
\multirow{4}{*}{}{10}
& FSDetView \cite{xiao2022few} & 1.3 $\pm$ 0.3 & 8.7 $\pm$ 2.1 & 4.6 $\pm$ 1.2 & 33.8 $\pm$ 0.5 & 29.8 $\pm$ 1.6 & 32.9 $\pm$ 3.4 \\
& TFA \cite{wang2020frustratingly} & 3.3 $\pm$ 0.8 & 9.0 $\pm$ 2.6 & 3.8 $\pm$ 1.1 & 58.6 $\pm$ 0.3 & 56.5 $\pm$ 0.8 & 59.0 $\pm$ 1.5\\
& DH-FSDet \cite{wolf2021double} & 5.2 $\pm$ 0.8 & 14.5 $\pm$ 1.7 & 9.7 $\pm$ 2.2 & \bf{65.0} $\pm$ 0.2 & \bf{64.5} $\pm$ 0.1 & \bf{67.8} $\pm$ 0.1\\
& Ours & \bf{10.2} $\pm$ 3.3 & \bf{17.7} $\pm$ 3.8 & \bf{14.0} $\pm$ 2.1 & 63.7 $\pm$ 0.4 & 62.4 $\pm$ 0.4 & 66.1 $\pm$ 0.7    \\ 
\addlinespace[0.5ex]
\cline{1-8}
\addlinespace[0.5ex]
\multirow{3}{*}{}
{50}
& FSDetView \cite{xiao2022few} & 7.2 $\pm$ 2.3 & 26.8 $\pm$ 2.8 & 17.1 $\pm$ 1.1 & 35.3 $\pm$ 0.5 & 30.0 $\pm$ 1.1 & 34.6 $\pm$ 1.1 \\
& TFA \cite{wang2020frustratingly} & 4.7 $\pm$ 0.0 & 12.1 $\pm$ 1.9 & 5.6 $\pm$ 1.4 & 60.7 $\pm$ 0.5 & 58.5 $\pm$ 0.8 & 60.9 $\pm$ 0.3 \\
& DH-FSDet \cite{wolf2021double}  & 12.8 $\pm$ 0.8 & 28.9 $\pm$ 3.4 & 19.6 $\pm$ 2.4 & \bf{65.1} $\pm$ 0.1 & \bf{64.7} $\pm$ 0.1 & \bf{68.0} $\pm$ 0.1\\
& Ours & \bf{24.8} $\pm$ 2.1 & \bf{39.3} $\pm$ 2.1 & \bf{31.1} $\pm$ 1.2 & 62.8 $\pm$ 0.4 & 62.6 $\pm$ 0.3 & 65.8 $\pm$ 0.1 \\
\addlinespace[0.5ex]
\cline{1-8}
\addlinespace[0.5ex]
\multirow{3}{*}{}
{100} 
& FSDetView \cite{xiao2022few} & 10.2 $\pm$ 1.2 & 32.8 $\pm$ 2.0 & 24.1 $\pm$ 1.1 & 36.4 $\pm$ 0.6 & 30.4 $\pm$ 0.4 & 34.5 $\pm$ 1.3 \\
& TFA \cite{wang2020frustratingly} & 5.0 $\pm$ 0.3 & 14.4 $\pm$ 1.5 & 5.4 $\pm$ 1.1 & 61.4 $\pm$ 0.3 & 59.2 $\pm$ 0.2 & 61.6 $\pm$ 0.4 \\
& DH-FSDet \cite{wolf2021double} & 16.7 $\pm$ 1.7 & 36.0 $\pm$ 1.7 & 23.1 $\pm$ 0.9 & \bf{65.2} $\pm$ 0.1 & \bf{64.8} $\pm$ 0.1 & \bf{68.1} $\pm$ 0.1 \\
& Ours & \bf{34.3} $\pm$ 1.9 & \bf{45.0} $\pm$ 1.0 & \bf{33.0} $\pm$ 1.3 & 63.3 $\pm$ 0.3 & 62.9 $\pm$ 0.2 & 65.6 $\pm$ 0.1 \\
\bottomrule
\end{tabular}
\end{center}
\label{tab:isaid}
\end{table*}

\begin{figure*}[htp]
    \centering
    \includegraphics[width=1.0\linewidth]{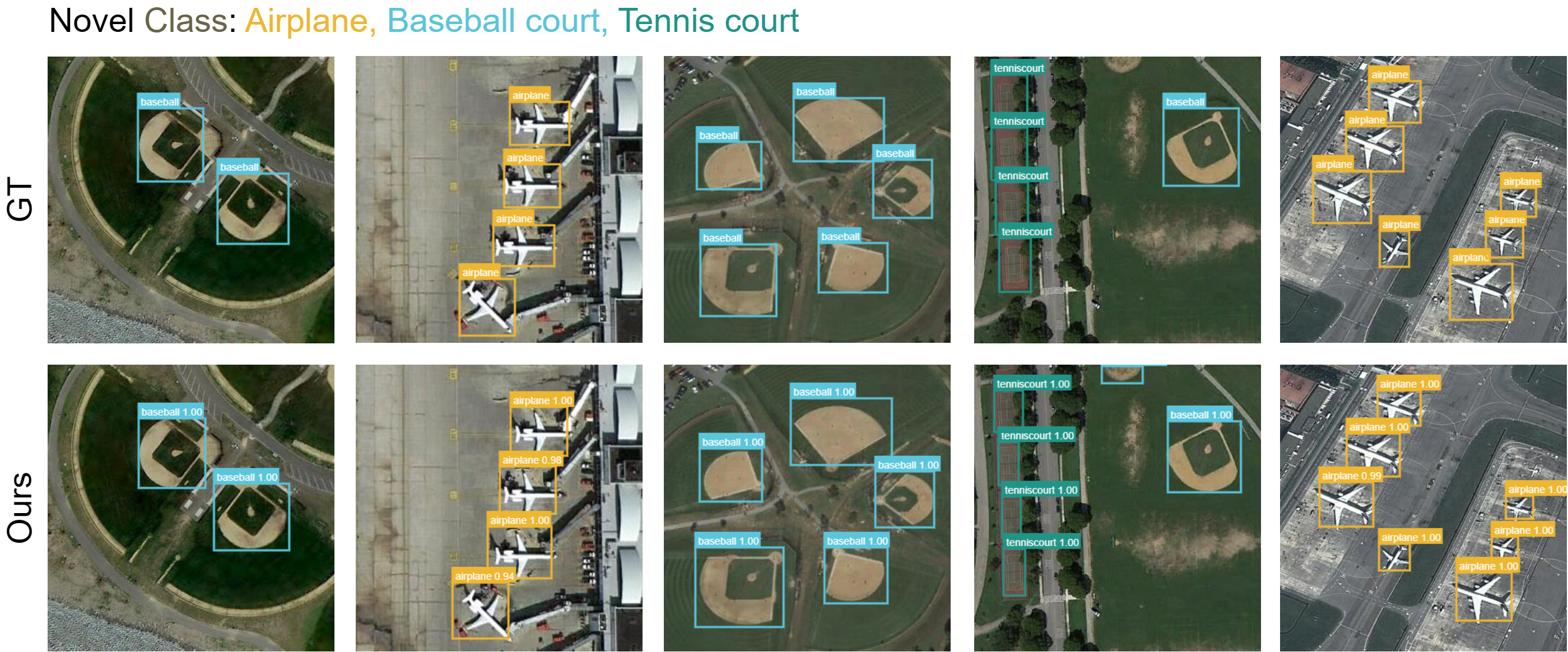}
    \caption{
        Visualized FSOD results of the proposed methods under $K=20$ shots setting on NWPU-VHR10 v2 dataset. The base-novel class split follows \cite{zhang2021oriented}.
    }
    \label{fig:nwpu}
\end{figure*}

\begin{figure*}[htp]
    \centering
    \includegraphics[width=1.0\linewidth]{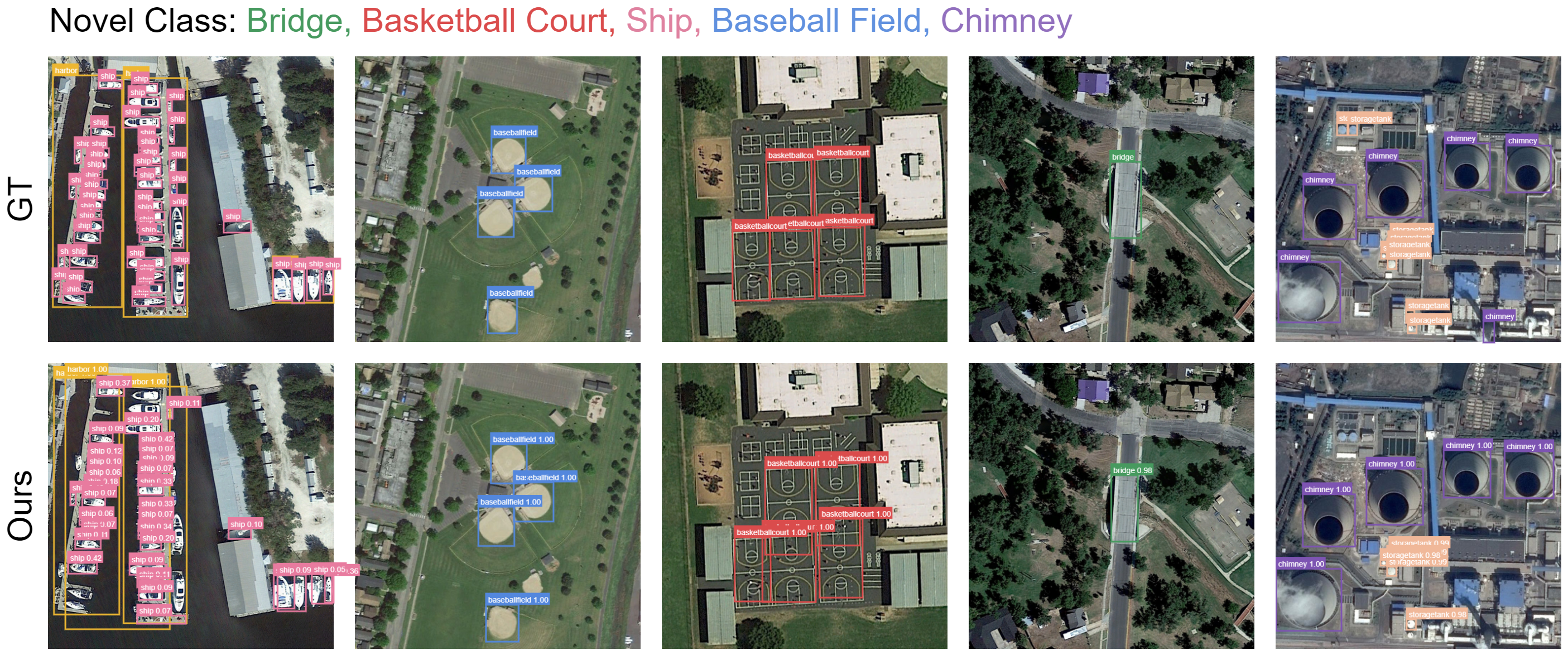}
    \caption{
        Visualized FSOD results of the proposed methods under $K=20$ shots setting on DIOR dataset. \FH{The base-novel class split follows the split 2 in \cite{cheng2021prototype}}.
    }
    \label{fig:dior_part1}
\end{figure*}

\begin{figure*}[htp]
    \centering
    \includegraphics[width=1.0\linewidth]{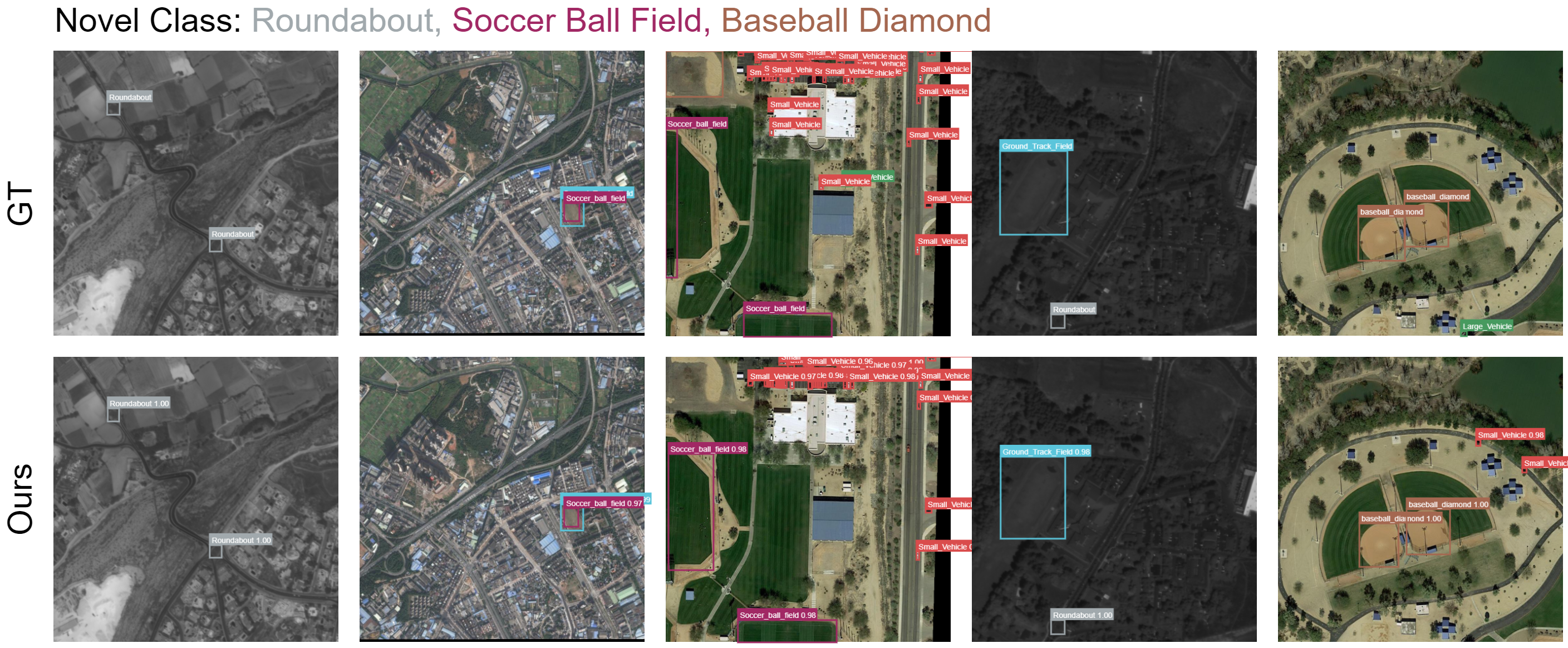}
    \caption{
        Visualized FSOD results of the proposed methods under $K=100$ shots setting on iSAID dataset. 
        \FH{The base-novel class split follows the split 2 in \cite{wolf2021double}.}
    }
    \label{fig:isaid_part1}
\end{figure*}

\begin{figure*}[htp]
    \centering
    \includegraphics[width=1.0\linewidth]{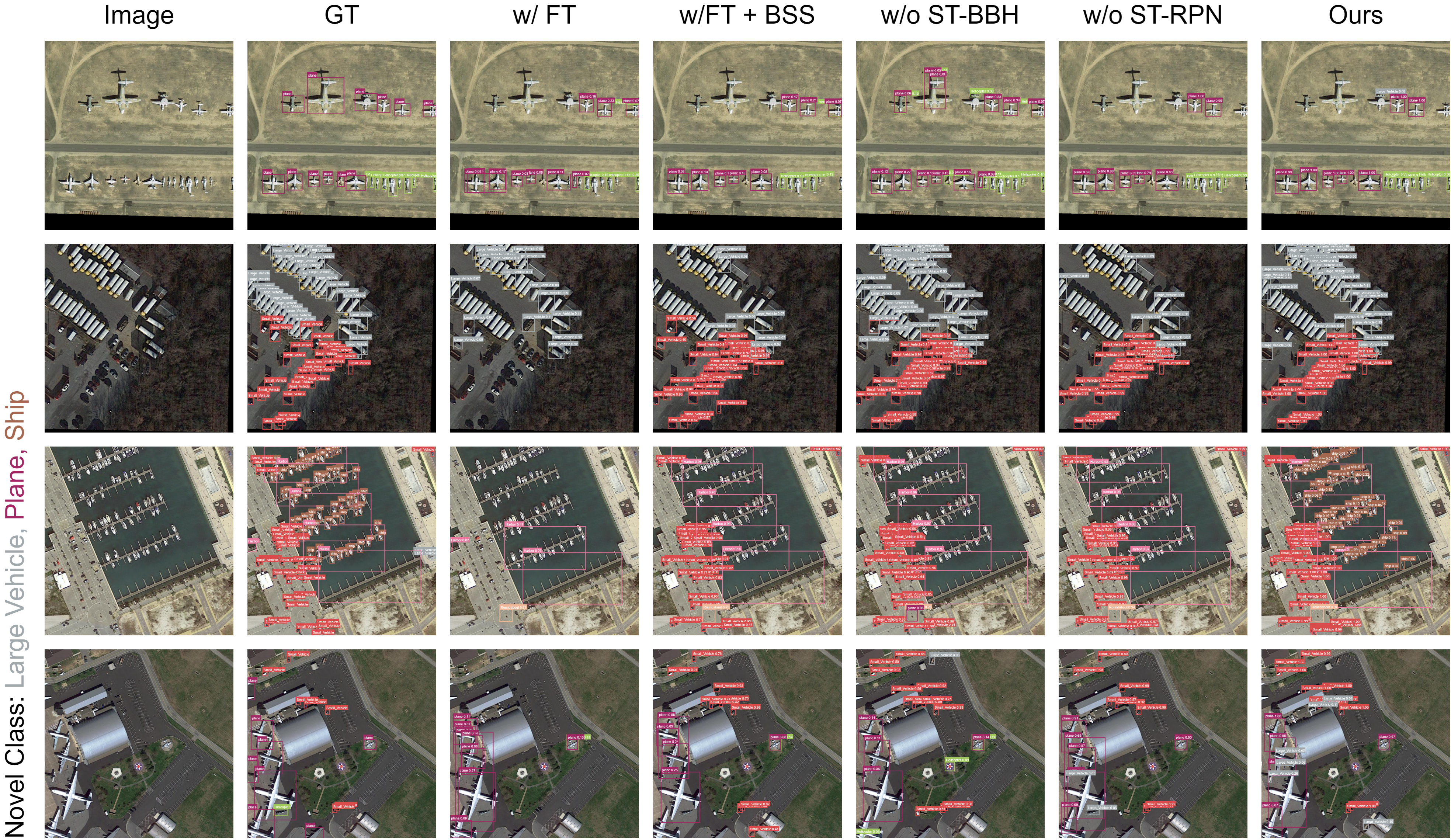}
    \caption{
        Visualized FSOD results of the proposed methods under different ablations on iSAID dataset.
        The number of novel shots is set to $K=100$. The base-novel class split is the split 1 in \cite{wolf2021double}.
        }
    \label{fig:isaid_compare}
\end{figure*}

\subsection{Quantitative Results}
\FH{
The quantitative results for the NWPU-VHR10 v2, DIOR, and iSAID datasets are presented in Table \ref{tab:nwpu}, Table \ref{tab:dior_part1} and \ref{tab:dior_part2}, and Table \ref{tab:isaid}, respectively.
Overall, our proposed method achieves superior or comparable performance across all datasets and various settings related to novel objects. Specifically:
}

\FH{
On the NWPU-VHR10 v2 dataset, our method attains state-of-the-art performance in both 3 and 20-shot settings, surpassing the second-best results by a notable margin ranging from 3\% to 7\%. Additionally, it performs comparably to state-of-the-art models in the 5 and 10-shot settings.}

\FH{
Regarding the DIOR dataset, as shown in Table \ref{tab:dior_part1}, our method consistently outperforms the second-best approach by a substantial margin of 5\% to 7\%. In Table \ref{tab:dior_part2}, our method excels in all settings except for the 20-shot setting of split 3. Remarkably, our results in split 1 demonstrate improvements exceeding 10\%.}

\FH{
On the iSAID dataset, our proposed method delivers significant enhancements, ranging from 5\% to over 10\%. These improvements remain consistent across various splits and different numbers of shots.}

\FH{
These results not only demonstrate the effectiveness of our proposed method but also underscore the significance of addressing the issue of incompletely annotated novel objects.
It is worth noting that performance variance tends to be higher in cases with fewer shots, suggesting that FSOD performance is somewhat sensitive to the sampling of annotated novel objects, especially when the available shots are limited.
}

\subsection{Qualitative Results}
\FH{
We present the visualization results on NWPU-VHR10 v2, DIOR, and iSAID datasets in Fig. \ref{fig:nwpu}, \ref{fig:dior_part1}, and \ref{fig:isaid_part1}.
Some observations and conclusions can be made from the results:
}

\begin{itemize}
    \item 
    \FH{
    The proposed method plays a crucial role in alleviating the challenge posed by incompletely annotated novel objects (IANO). 
    This issue becomes particularly pronounced in scenarios featuring multiple small objects within a single image.
    For instance, in the first column of Fig. \ref{fig:dior_part1}, ships can be easily mistaken as part of the background by a few-shot object detector. This occurs because, during the few-shot fine-tuning stage, a significant number of ship annotations will be missing, as depicted in Fig. \ref{fig:motivation}.
    In contrast to this, our detection results show demonstrate that our method excels in recalling many ship objects.
    This verify the effectiveness of employing self-training technique to solve the IANO issue.}

    \item     
    \FH{
    By separating the proposals for base and novel class objects, our approach effectively preserves the performance of the few-shot detector on the base class.
    As elaborated in Section \ref{sec:method_st_rpn}, this separation of region proposals serves to prevent the few-shot fine-tuning process from negatively influence the detection of base class objects.
    As evidenced by the results depicted in the 5th column of Fig. \ref{fig:dior_part1} and the 3rd column of Fig. \ref{fig:isaid_part1},
    our method is able to detect challenging, small base class objects such as storage tanks and vehicles.}
\end{itemize}

\setlength\tabcolsep{0pt}
\begin{table*}
\footnotesize
\caption{
    Ablation study results on split1 of the iSAID benchmark.
    The same seed for image sampling and other randomized process is used to make a fair comparison.
    ``w/ FT'' denotes whether the few-shot fine-tuning \FH{approach proposed in \cite{wang2020frustratingly}} is applied.
    ``BSS'' denotes the balanced sampling strategy as described in Sec. \ref{sec:method_bss}.
}
\renewcommand\arraystretch{1.2}
\newcolumntype{C}{>{\centering \arraybackslash}m{1.5cm}}
\newcolumntype{D}{>{\centering \arraybackslash}m{1.5cm}}
\begin{center}
    \begin{tabular}{C C C C | C C C | C C C}
\toprule
        \multirow{2}{*}{\parbox{0.1\linewidth}{\vspace{0.1cm}}
        w/ FT} &
        \multirow{2}{*}{\parbox{0.1\linewidth}{\vspace{0.1cm}}
        BSS} &
        \multirow{2}{*}{\parbox{0.1\linewidth}{\vspace{0.1cm}}
        ST-RPN } &
        \multirow{2}{*}{\parbox{0.1\linewidth}{\vspace{0.1cm}}
        ST-BBH }
         & \multicolumn{3}{c|}{Novel Classes} & \multicolumn{3}{c}{Base Classes} \\
 & & & & 10 & 50 & 100 & 10 & 50 & 100  \\

\addlinespace[0.5ex]
\cline{1-10}
\addlinespace[0.5ex]
& & & & - & - & - & 65.8 & 65.8 & 65.8 \\
\checkmark & & & & 3.7 & 13.1 & 17.0 & 39.6 & 54.2 & 57.8\\
\checkmark & \checkmark & & & 4.0 & 12.3 & 16.2 & \bf{65.9} & 65.9 & 66.0  \\
\checkmark & \checkmark & \checkmark & & 5.2 & 16.3 & 22.3 & 65.5 & \bf{66.0} & \bf{66.4} \\
\checkmark & \checkmark & & \checkmark & 6.5 & 17.8 & 27.0 & 65.1 & 64.9 & 65.0 \\
\checkmark & \checkmark & \checkmark & \checkmark & \bf{10.3} & \bf{23.4} & \bf{35.5} & 63.0 & 62.3 & 63.0\\

\bottomrule
\end{tabular}
\end{center}

\label{tab:ablation}
\end{table*}

\setlength\tabcolsep{0pt}
\begin{table*}
\footnotesize
\caption{
    Sensitivity analyses of the hyper-parameters on split1 of the iSAID benchmark.
}
\renewcommand\arraystretch{1.2}
\newcolumntype{C}{>{\centering \arraybackslash}m{1.5cm}}
\newcolumntype{D}{>{\centering \arraybackslash}m{1.5cm}}
\begin{center}
    \begin{tabular}{C C C | C C C | C C C}
\toprule
        \multirow{2}{*}{\parbox{0.1\linewidth}{\vspace{0.1cm}}
        $\tau_{\text{rpn}}$} &
        \multirow{2}{*}{\parbox{0.1\linewidth}{\vspace{0.1cm}}
        $\tau_{\text{bbh}}$} &
        \multirow{2}{*}{\parbox{0.1\linewidth}{\vspace{0.1cm}}
        $\alpha$ }
         & \multicolumn{3}{c|}{Novel Classes} & \multicolumn{3}{c}{Base Classes} \\
 & & & 10 & 50 & 100 & 10 & 50 & 100  \\

\addlinespace[0.5ex]
\cline{1-9}
\addlinespace[0.5ex]
0.8 & 0.8 & 0.999 & 10.3 & 23.4 & 35.5 & 63.0 & 62.3 & 63.0 \\
0.8 & 0.6 & 0.999 & 10.1 & \bf{25.1} & 34.4 & 63.8 & \bf{63.2} & 62.9\\
0.8 & 0.9 & 0.999 & 10.0 & 22.8 & \bf{36.5} & \bf{64.0} & 62.4 & 62.9\\
0.6 & 0.8 & 0.999 & \bf{10.5} & 22.4 & 35.0 & 63.4 & 62.4 & 62.5\textbf{}\\
0.9 & 0.8 & 0.999 & 10.5 & 23.7 & 36.4 & 63.9 & 61.9 & \bf{63.3}\\
0.8 & 0.8 & 0.99 & 10.4 & 23.9 & 35.3 & 63.5 & 62.7 & \bf{63.3} \\
\bottomrule
\end{tabular}
\end{center}

\label{tab:sensitivity}
\end{table*}

\setlength\tabcolsep{0pt}
\begin{table}
\footnotesize
\caption{
    \FH{Training times (seconds per iteration) and inference frames per second (FPS) of the proposed method evaluated on iSAID dataset. mAP (\%) is also reported for comparison.}
}
\renewcommand\arraystretch{1.2}
\newcolumntype{C}{>{\centering \arraybackslash}m{2.2cm}}
\newcolumntype{D}{>{\centering \arraybackslash}m{1.4cm}}
\begin{center}
    \begin{tabular}{D D | D | C | C }
\toprule
        ST-RPN & ST-BBH & mAP
        & Training Times
        & Inference FPS \\
\addlinespace[0.5ex]
\cline{1-5}
\addlinespace[0.5ex]
 & & 16.2 & 0.258 & 18.8 \\
 \checkmark & & 22.3 & 0.458 & 16.0 \\
  & \checkmark & 27.0 & 0.260 & 18.7 \\
 \checkmark & \checkmark & 35.5 & 0.479 & 15.9 \\

\bottomrule
\end{tabular}
\end{center}

\label{tab:efficiency}
\end{table}

\subsection{Ablation Studies}
\label{sec:expr_ablation}
We conduct ablation studies on the first split of iSAID dataset to better understand the effect of all the components used in the proposed method. The results are presented in Table \ref{tab:ablation} and Fig. \ref{fig:isaid_compare} for quantitative and qualitative analysis, respectively. The following observations can be made from the results:
\begin{itemize}

    \item
    \FH{
    The naive fine-tuning strategy introduced in \cite{wang2020frustratingly} has a detrimental effect on the detection of base class objects. A noticeable performance decline becomes evident when comparing the results in the first row to those in the second row, according to Table \ref{tab:ablation}.
    This decline in performance stems from the fact that this particular strategy employs only a limited subset of base class annotations for model fine-tuning, leading to incomplete annotations for the base class.
    }
    
    \item 
    \FH{
    BSS proves to be highly effective in preserving the base class performance after the few-shot fine-tuning process, at the cost of a slight trade-off in novel class performance. 
    This highlights the critical importance of fine-tuning with a complete base annotation set for the maintenance of base class performance in satellite image-based FSOD.}
    
    \item
    Applying the proposed ST-RPN module does not influence the base class performance, owing to the separation of the extraction of base and novel proposals. In addition, applying the proposed ST-RPN module is beneficial to improving the performance on novel class.
    
    \item  
    After applying the proposed ST-BBH module, there is a slight performance decay on the base classes. One possible reason is that since the base and novel classes share the same bounding box head (BBH), the fine-tuning on novel objects affects the general feature extraction within BBH and further affects the detection of base classes. However, there is a consistent improvement on the novel classes after applying ST-BBH, ranging from 3\% to even 10\%, at different number of shots.

    \item
    By combining all the modules, the highest novel class accuracy is achieved, with a margin of 6\% to nearly 20\% compared to the naive fine-tuning strategy. This verifies the effectiveness of the proposed ST-RPN and ST-BBH modules in solving the IANO issue.
\end{itemize}

Fig. \ref{fig:isaid_compare} shows visualized results for different ablated models. The second row shows that using the ST-RPN module helps to recall more large vehicle objects. However, using ST-RPN alone without ST-BBH may lead to a higher false positive rate, as can be seen from the first row. By combining the ST-RPN and ST-BBH, the best visualization quality is achieved. The third row demonstrates that this combination can help to detect small ship objects with high accuracy. These results further demonstrate the significance of incorporating the self-training mechanism to solve the unlabeled novel object issue.

\begin{figure*}[htp]
    \centering
    \includegraphics[width=1.0\linewidth]{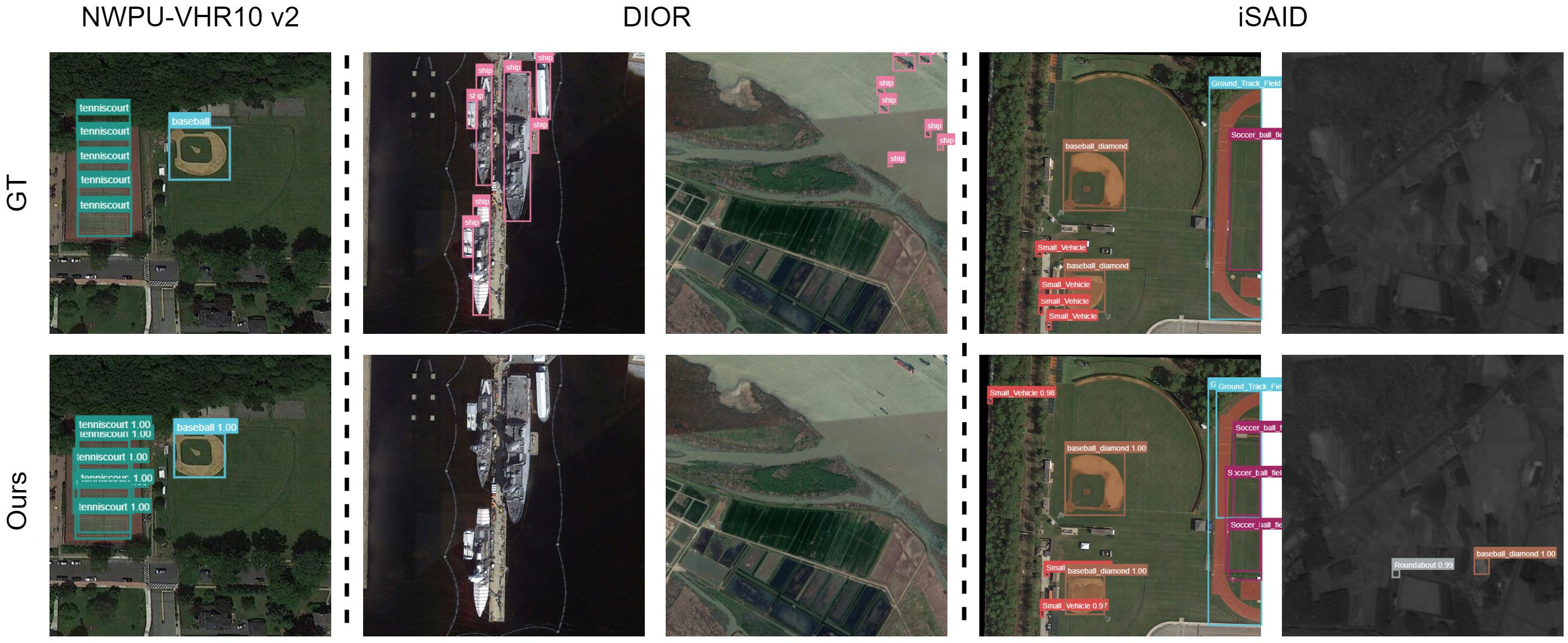}
    \caption{
        \FH{Visualized failure cases of the proposed method on NWPU-VHR10 v2, DIOR and iSAID datasets. Base-novel class splits are the same as the splits in Fig. \ref{fig:nwpu}, \ref{fig:dior_part1}, and \ref{fig:isaid_part1}.}
    }
    \label{fig:failure_cases}
\end{figure*}

\subsection{Sensitivity Analyses of Hyperparameters}
\label{sec:expr_parameter}
We conducted a sensitivity analysis to evaluate the impact of hyper-parameter selection on the proposed method. The hyper-parameters we tested include the two self-training thresholds $\tau_{\text{rpn}}$ and $\tau_{\text{bbh}}$ used in ST-RPN and ST-BBH, and the momentum $\alpha$ used when updating the teacher networks via EMA \cite{araslanov2021self}. The results are presented in Tab. \ref{tab:sensitivity}. We observed slight performance fluctuations (generally less than $2\%$) with different hyper-parameter values. However, compared to the variances caused by different sampling seeds as shown in Tab. \ref{tab:dior_part1}, \ref{tab:dior_part2} and \ref{tab:isaid}, such fluctuations are not significant. Therefore, we can conclude that the proposed method is not highly sensitive to the values of the aforementioned hyper-parameters.

\FH{
\subsection{Computational Efficiency}
To assess the incremental computational demands introduced by our two proposed modules, ST-RPN and ST-BBH, we conducted an evaluation based on training times (measured in seconds per iteration) during the fine-tuning stage and inference frames per second (FPS) when these modules were integrated.
The outcomes, as illustrated in Table \ref{tab:efficiency}, indicate that the inclusion of ST-BBH results in a negligible computational overhead, both during training and inference phases. This outcome aligns with our expectations, as ST-BBH simply introduces a pair of additional bounding box regressor and classifier layers on top of the Region of Interest (RoI) features.
}

\FH{In contrast, the integration of ST-RPN introduces a more substantial computational load during the training phase. 
However, it is crucial to note that the fine-tuning phase in FSOD algorithms typically involves significantly fewer iterations compared to the base training phase (see our settings in Section \ref{sec:expr_details}). 
As such, the additional training costs remain within acceptable bounds.
Furthermore, while ST-RPN does marginally reduce inference speed, it is essential to consider this cost in light of the performance gains it offers. This expense is manageable given the corresponding boost in detection performance.
Additionally, upon closer investigation, we found that a substantial portion of the computational cost introduced by ST-RPN is attributed to the additional bounding box operations, such as non-maximum suppression (NMS).
These options can be optimized and accelerated through the utilization of CUDA implementations.
}

\subsection{Failure Cases}
To gain a better understanding of the limitations of the proposed FSOD method, we visualize some failure cases in different FSOD settings, as shown in Fig. \ref{fig:failure_cases}.
Based on the figure, we can observe that the majority of the missed detections are primarily due to the small size of the objects \FH{(``ships'' in the 3rd column)}, or objects with large size variance compared to the training data \FH{(``ships'' in the second column)}. Additionally, there may be duplicated detected boxes for objects that lack clear boundaries, such as the \FH{``tennis court'' in the 1st column or ``soccerball field'' in the 4th column}.
While these issues are prevalent in general object detection \cite{rabbi2020small}, improved techniques related to addressing them can also be utilized to enhance the performance of FSOD.

\section{Conclusion}
In this paper, we analyzed the current FSOD setting for remote sensing and identified the issue of \FH{incompletely annotated novel objects} that can negatively impact the performance of FSOD methods. To address this issue, we propose to incorporate the self-training mechanism into the classical two-stage fine-tuning based FSOD pipeline. Our approach includes a ST-RPN module, which generates a set of novel proposals by excluding some proposals from the loss calculation that are likely to be novel objects but cannot be assigned to an existing few-shot annotation. Additionally, we designed a ST-BBH module that leverages the pseudo labels generated from a teacher BBH to filter out potential novel bounding boxes that are unlabeled and use them to supervise the student BBH to recall more novel objects.

While our proposed method significantly improved the novel class FSOD performance in remote sensing, the base class performance may slightly decrease compared to the base model. Future works could focus on designing a generalized FSOD method that prevents the model from forgetting the previously learned base knowledge while improving the performance in detecting novel classes.


\bibliographystyle{IEEEtran}
\bibliography{mybib}

\begin{thebibliography}{10}
\providecommand{\url}[1]{#1}
\csname url@samestyle\endcsname
\providecommand{\newblock}{\relax}
\providecommand{\bibinfo}[2]{#2}
\providecommand{\BIBentrySTDinterwordspacing}{\spaceskip=0pt\relax}
\providecommand{\BIBentryALTinterwordstretchfactor}{4}
\providecommand{\BIBentryALTinterwordspacing}{\spaceskip=\fontdimen2\font plus
\BIBentryALTinterwordstretchfactor\fontdimen3\font minus
  \fontdimen4\font\relax}
\providecommand{\BIBforeignlanguage}[2]{{%
\expandafter\ifx\csname l@#1\endcsname\relax
\typeout{** WARNING: IEEEtran.bst: No hyphenation pattern has been}%
\typeout{** loaded for the language `#1'. Using the pattern for}%
\typeout{** the default language instead.}%
\else
\language=\csname l@#1\endcsname
\fi
#2}}
\providecommand{\BIBdecl}{\relax}
\BIBdecl

\bibitem{huang2022survey}
G.~Huang, I.~Laradji, D.~Vazquez, S.~Lacoste-Julien, and P.~Rodriguez, ``A
  survey of self-supervised and few-shot object detection,'' \emph{IEEE
  Transactions on Pattern Analysis and Machine Intelligence}, vol.~45, no.~04,
  pp. 4071--4089, 2023.

\bibitem{ren2015faster}
S.~Ren, K.~He, R.~Girshick, and J.~Sun, ``Faster r-cnn: Towards real-time
  object detection with region proposal networks,'' \emph{Advances in neural
  information processing systems}, vol.~28, 2015.

\bibitem{redmon2016you}
J.~Redmon, S.~Divvala, R.~Girshick, and A.~Farhadi, ``You only look once:
  Unified, real-time object detection,'' in \emph{Proceedings of the IEEE
  conference on computer vision and pattern recognition}, 2016, pp. 779--788.

\bibitem{carion2020end}
N.~Carion, F.~Massa, G.~Synnaeve, N.~Usunier, A.~Kirillov, and S.~Zagoruyko,
  ``End-to-end object detection with transformers,'' in \emph{Computer
  Vision--ECCV 2020: 16th European Conference, Glasgow, UK, August 23--28,
  2020, Proceedings, Part I 16}.\hskip 1em plus 0.5em minus 0.4em\relax
  Springer, 2020, pp. 213--229.

\bibitem{5206848}
J.~Deng, W.~Dong, R.~Socher, L.-J. Li, K.~Li, and L.~Fei-Fei, ``{ImageNet}: A
  large-scale hierarchical image database,'' in \emph{2009 IEEE Conference on
  Computer Vision and Pattern Recognition}, 2009, pp. 248--255.

\bibitem{lin2014microsoft}
T.-Y. Lin, M.~Maire, S.~Belongie, J.~Hays, P.~Perona, D.~Ramanan,
  P.~Doll{\'a}r, and C.~L. Zitnick, ``Microsoft coco: Common objects in
  context,'' in \emph{Computer Vision--ECCV 2014: 13th European Conference,
  Zurich, Switzerland, September 6-12, 2014, Proceedings, Part V 13}.\hskip 1em
  plus 0.5em minus 0.4em\relax Springer, 2014, pp. 740--755.

\bibitem{10075555}
J.~Zhang, J.~Lei, W.~Xie, Z.~Fang, Y.~Li, and Q.~Du, ``{SuperYOLO}: Super
  resolution assisted object detection in multimodal remote sensing imagery,''
  \emph{IEEE Transactions on Geoscience and Remote Sensing}, vol.~61, pp.
  1--15, 2023.

\bibitem{dai2022ao2}
L.~Dai, H.~Liu, H.~Tang, Z.~Wu, and P.~Song, ``Ao2-detr: Arbitrary-oriented
  object detection transformer,'' \emph{IEEE Transactions on Circuits and
  Systems for Video Technology}, 2022.

\bibitem{zhang2023semi}
X.~Zhang, Y.~Feng, S.~Zhang, N.~Wang, S.~Mei, and M.~He, ``Semi-supervised
  person detection in aerial images with instance segmentation and maximum mean
  discrepancy distance,'' \emph{Remote Sensing}, vol.~15, no.~11, p. 2928,
  2023.

\bibitem{wang2020generalizing}
Y.~Wang, Q.~Yao, J.~T. Kwok, and L.~M. Ni, ``Generalizing from a few examples:
  A survey on few-shot learning,'' \emph{ACM computing surveys (csur)},
  vol.~53, no.~3, pp. 1--34, 2020.

\bibitem{lang2022learning}
C.~Lang, G.~Cheng, B.~Tu, and J.~Han, ``Learning what not to segment: A new
  perspective on few-shot segmentation,'' in \emph{Proceedings of the IEEE/CVF
  conference on computer vision and pattern recognition}, 2022, pp. 8057--8067.

\bibitem{xiong2022doubly}
Z.~Xiong, H.~Li, and X.~X. Zhu, ``Doubly deformable aggregation of covariance
  matrices for few-shot segmentation,'' in \emph{European Conference on
  Computer Vision}.\hskip 1em plus 0.5em minus 0.4em\relax Springer, 2022, pp.
  133--150.

\bibitem{vinyals2016matching}
O.~Vinyals, C.~Blundell, T.~Lillicrap, D.~Wierstra \emph{et~al.}, ``Matching
  networks for one shot learning,'' \emph{Advances in neural information
  processing systems}, vol.~29, 2016.

\bibitem{cheng2022holistic}
G.~Cheng, C.~Lang, and J.~Han, ``Holistic prototype activation for few-shot
  segmentation,'' \emph{IEEE Transactions on Pattern Analysis and Machine
  Intelligence}, vol.~45, no.~4, pp. 4650--4666, 2022.

\bibitem{lang2023base}
C.~Lang, G.~Cheng, B.~Tu, C.~Li, and J.~Han, ``Base and meta: A new perspective
  on few-shot segmentation,'' \emph{IEEE Transactions on Pattern Analysis and
  Machine Intelligence}, 2023.

\bibitem{chen2018lstd}
H.~Chen, Y.~Wang, G.~Wang, and Y.~Qiao, ``{LSTD}: A low-shot transfer detector
  for object detection,'' in \emph{Proceedings of the AAAI conference on
  artificial intelligence}, vol.~32, no.~1, 2018.

\bibitem{wang2020frustratingly}
X.~Wang, T.~E. Huang, T.~Darrell, J.~E. Gonzalez, and F.~Yu, ``Frustratingly
  simple few-shot object detection,'' \emph{arXiv preprint arXiv:2003.06957},
  2020.

\bibitem{9884721}
L.~Wang, S.~Zhang, Z.~Han, Y.~Feng, J.~Wei, and S.~Mei, ``Diversity
  measurement-based meta-learning for few-shot object detection of remote
  sensing images,'' in \emph{IGARSS 2022 - 2022 IEEE International Geoscience
  and Remote Sensing Symposium}, 2022, pp. 3087--3090.

\bibitem{qiao2023few}
B.~Qiao, H.~Zhou, L.~Yang, and X.~Xie, ``Few shot object detection with
  incompletely annotated samples,'' in \emph{2023 International Joint
  Conference on Neural Networks (IJCNN)}.\hskip 1em plus 0.5em minus
  0.4em\relax IEEE, 2023, pp. 1--7.

\bibitem{li2021few}
X.~Li, J.~Deng, and Y.~Fang, ``Few-shot object detection on remote sensing
  images,'' \emph{IEEE Transactions on Geoscience and Remote Sensing}, vol.~60,
  pp. 1--14, 2021.

\bibitem{waqas2019isaid}
S.~Waqas~Zamir, A.~Arora, A.~Gupta, S.~Khan, G.~Sun, F.~Shahbaz~Khan, F.~Zhu,
  L.~Shao, G.-S. Xia, and X.~Bai, ``{iSAID}: A large-scale dataset for instance
  segmentation in aerial images,'' in \emph{Proceedings of the IEEE/CVF
  Conference on Computer Vision and Pattern Recognition Workshops}, 2019, pp.
  28--37.

\bibitem{farahani2021brief}
A.~Farahani, S.~Voghoei, K.~Rasheed, and H.~R. Arabnia, ``A brief review of
  domain adaptation,'' \emph{Advances in data science and information
  engineering}, pp. 877--894, 2021.

\bibitem{wurm2019semantic}
M.~Wurm, T.~Stark, X.~X. Zhu, M.~Weigand, and H.~Taubenb{\"o}ck, ``Semantic
  segmentation of slums in satellite images using transfer learning on fully
  convolutional neural networks,'' \emph{ISPRS journal of photogrammetry and
  remote sensing}, vol. 150, pp. 59--69, 2019.

\bibitem{qingsong2023uni}
Q.~Xu, Y.~Shi, X.~Yuan, and X.~Zhu, ``Universal domain adaptation for remote
  sensing image scene classification,'' \emph{IEEE Transactions on Geoscience
  and Remote Sensing}, vol.~61, pp. 1--15, 01 2023.

\bibitem{wei2021crest}
C.~Wei, K.~Sohn, C.~Mellina, A.~Yuille, and F.~Yang, ``Crest: A
  class-rebalancing self-training framework for imbalanced semi-supervised
  learning,'' in \emph{Proceedings of the IEEE/CVF Conference on Computer
  Vision and Pattern Recognition}, 2021, pp. 10\,857--10\,866.

\bibitem{girshick2015fast}
R.~Girshick, ``Fast r-cnn,'' in \emph{Proceedings of the IEEE international
  conference on computer vision}, 2015, pp. 1440--1448.

\bibitem{araslanov2021self}
N.~Araslanov and S.~Roth, ``Self-supervised augmentation consistency for
  adapting semantic segmentation,'' in \emph{Proceedings of the IEEE/CVF
  Conference on Computer Vision and Pattern Recognition}, 2021, pp.
  15\,384--15\,394.

\bibitem{viola2001rapid}
P.~Viola and M.~Jones, ``Rapid object detection using a boosted cascade of
  simple features,'' in \emph{Proceedings of the 2001 IEEE computer society
  conference on computer vision and pattern recognition. CVPR 2001},
  vol.~1.\hskip 1em plus 0.5em minus 0.4em\relax Ieee, 2001, pp. I--I.

\bibitem{thanh2010improved}
N.~D. Thanh, W.~Li, and P.~Ogunbona, ``An improved template matching method for
  object detection,'' in \emph{Computer Vision--ACCV 2009: 9th Asian Conference
  on Computer Vision, Xi’an, September 23-27, 2009, Revised Selected Papers,
  Part III 9}.\hskip 1em plus 0.5em minus 0.4em\relax Springer, 2010, pp.
  193--202.

\bibitem{yang2022efficient}
Z.~Yang, C.~Zhang, R.~Li, Y.~Xu, and G.~Lin, ``Efficient few-shot object
  detection via knowledge inheritance,'' \emph{IEEE Transactions on Image
  Processing}, vol.~32, pp. 321--334, 2022.

\bibitem{karlinsky2019repmet}
L.~Karlinsky, J.~Shtok, S.~Harary, E.~Schwartz, A.~Aides, R.~Feris, R.~Giryes,
  and A.~M. Bronstein, ``Repmet: Representative-based metric learning for
  classification and few-shot object detection,'' in \emph{Proceedings of the
  IEEE/CVF conference on computer vision and pattern recognition}, 2019, pp.
  5197--5206.

\bibitem{fan2020few}
Q.~Fan, W.~Zhuo, C.-K. Tang, and Y.-W. Tai, ``Few-shot object detection with
  attention-rpn and multi-relation detector,'' in \emph{Proceedings of the
  IEEE/CVF conference on computer vision and pattern recognition}, 2020, pp.
  4013--4022.

\bibitem{kang2019few}
B.~Kang, Z.~Liu, X.~Wang, F.~Yu, J.~Feng, and T.~Darrell, ``Few-shot object
  detection via feature reweighting,'' in \emph{Proceedings of the IEEE/CVF
  International Conference on Computer Vision}, 2019, pp. 8420--8429.

\bibitem{xiao2022few}
Y.~Xiao, V.~Lepetit, and R.~Marlet, ``Few-shot object detection and viewpoint
  estimation for objects in the wild,'' \emph{IEEE Transactions on Pattern
  Analysis and Machine Intelligence}, vol.~45, no.~3, pp. 3090--3106, 2022.

\bibitem{huang2021few}
X.~Huang, B.~He, M.~Tong, D.~Wang, and C.~He, ``Few-shot object detection on
  remote sensing images via shared attention module and balanced fine-tuning
  strategy,'' \emph{Remote Sensing}, vol.~13, no.~19, p. 3816, 2021.

\bibitem{wolf2021double}
S.~Wolf, J.~Meier, L.~Sommer, and J.~Beyerer, ``Double head predictor based
  few-shot object detection for aerial imagery,'' in \emph{Proceedings of the
  IEEE/CVF International Conference on Computer Vision}, 2021, pp. 721--731.

\bibitem{zhang2023generalized}
T.~Zhang, X.~Zhang, P.~Zhu, X.~Jia, X.~Tang, and L.~Jiao, ``Generalized
  few-shot object detection in remote sensing images,'' \emph{ISPRS Journal of
  Photogrammetry and Remote Sensing}, vol. 195, pp. 353--364, 2023.

\bibitem{lu2023few}
X.~Lu, X.~Sun, W.~Diao, Y.~Mao, J.~Li, Y.~Zhang, P.~Wang, and K.~Fu, ``Few-shot
  object detection in aerial imagery guided by text-modal knowledge,''
  \emph{IEEE Transactions on Geoscience and Remote Sensing}, 2023.

\bibitem{zhang2023text}
S.~Zhang, F.~Song, X.~Liu, X.~Hao, Y.~Liu, T.~Lei, and P.~Jiang, ``Text
  semantic fusion relation graph reasoning for few-shot object detection on
  remote sensing images,'' \emph{Remote Sensing}, vol.~15, no.~5, p. 1187,
  2023.

\bibitem{yang2020fda}
Y.~Yang and S.~Soatto, ``{FDA}: Fourier domain adaptation for semantic
  segmentation,'' in \emph{Proceedings of the IEEE/CVF Conference on Computer
  Vision and Pattern Recognition}, 2020, pp. 4085--4095.

\bibitem{subhani2020learning}
M.~N. Subhani and M.~Ali, ``Learning from scale-invariant examples for domain
  adaptation in semantic segmentation,'' in \emph{Computer Vision--ECCV 2020:
  16th European Conference, Glasgow, UK, August 23--28, 2020, Proceedings, Part
  XXII 16}.\hskip 1em plus 0.5em minus 0.4em\relax Springer, 2020, pp.
  290--306.

\bibitem{mei2020instance}
K.~Mei, C.~Zhu, J.~Zou, and S.~Zhang, ``Instance adaptive self-training for
  unsupervised domain adaptation,'' in \emph{Computer Vision--ECCV 2020: 16th
  European Conference, Glasgow, UK, August 23--28, 2020, Proceedings, Part XXVI
  16}.\hskip 1em plus 0.5em minus 0.4em\relax Springer, 2020, pp. 415--430.

\bibitem{shin2020two}
I.~Shin, S.~Woo, F.~Pan, and I.~S. Kweon, ``Two-phase pseudo label
  densification for self-training based domain adaptation,'' in \emph{European
  conference on computer vision}.\hskip 1em plus 0.5em minus 0.4em\relax
  Springer, 2020, pp. 532--548.

\bibitem{10141556}
F.~Zhang, Y.~Shi, Z.~Xiong, W.~Huang, and X.~X. Zhu, ``Pseudo features guided
  self-training for domain adaptive semantic segmentation of satellite
  images,'' \emph{IEEE Transactions on Geoscience and Remote Sensing}, pp.
  1--1, 2023.

\bibitem{berthelot2019mixmatch}
D.~Berthelot, N.~Carlini, I.~Goodfellow, N.~Papernot, A.~Oliver, and C.~A.
  Raffel, ``{MixMatch}: A holistic approach to semi-supervised learning,''
  \emph{Advances in neural information processing systems}, vol.~32, 2019.

\bibitem{hoyer2022daformer}
L.~Hoyer, D.~Dai, and L.~Van~Gool, ``{DAFormer}: Improving network
  architectures and training strategies for domain-adaptive semantic
  segmentation,'' in \emph{Proceedings of the IEEE/CVF Conference on Computer
  Vision and Pattern Recognition}, 2022, pp. 9924--9935.

\bibitem{sohn2020fixmatch}
K.~Sohn, D.~Berthelot, N.~Carlini, Z.~Zhang, H.~Zhang, C.~A. Raffel, E.~D.
  Cubuk, A.~Kurakin, and C.-L. Li, ``{FixMatch}: Simplifying semi-supervised
  learning with consistency and confidence,'' \emph{Advances in neural
  information processing systems}, vol.~33, pp. 596--608, 2020.

\bibitem{li2021few_2}
Y.~Li, H.~Zhu, Y.~Cheng, W.~Wang, C.~S. Teo, C.~Xiang, P.~Vadakkepat, and T.~H.
  Lee, ``Few-shot object detection via classification refinement and distractor
  retreatment,'' in \emph{Proceedings of the IEEE/CVF Conference on Computer
  Vision and Pattern Recognition}, 2021, pp. 15\,395--15\,403.

\bibitem{lin2017feature}
T.-Y. Lin, P.~Doll{\'a}r, R.~Girshick, K.~He, B.~Hariharan, and S.~Belongie,
  ``Feature pyramid networks for object detection,'' in \emph{Proceedings of
  the IEEE conference on computer vision and pattern recognition}, 2017, pp.
  2117--2125.

\bibitem{srivastava2014dropout}
N.~Srivastava, G.~Hinton, A.~Krizhevsky, I.~Sutskever, and R.~Salakhutdinov,
  ``Dropout: a simple way to prevent neural networks from overfitting,''
  \emph{The journal of machine learning research}, vol.~15, no.~1, pp.
  1929--1958, 2014.

\bibitem{cheng2014multi}
G.~Cheng, J.~Han, P.~Zhou, and L.~Guo, ``Multi-class geospatial object
  detection and geographic image classification based on collection of part
  detectors,'' \emph{ISPRS Journal of Photogrammetry and Remote Sensing},
  vol.~98, pp. 119--132, 2014.

\bibitem{li2020object}
K.~Li, G.~Wan, G.~Cheng, L.~Meng, and J.~Han, ``Object detection in optical
  remote sensing images: A survey and a new benchmark,'' \emph{ISPRS journal of
  photogrammetry and remote sensing}, vol. 159, pp. 296--307, 2020.

\bibitem{zhang2021oriented}
Z.~Zhang, J.~Hao, C.~Pan, and G.~Ji, ``Oriented feature augmentation for
  few-shot object detection in remote sensing images,'' in \emph{2021 IEEE
  International Conference on Computer Science, Electronic Information
  Engineering and Intelligent Control Technology (CEI)}.\hskip 1em plus 0.5em
  minus 0.4em\relax IEEE, 2021, pp. 359--366.

\bibitem{cheng2021prototype}
G.~Cheng, B.~Yan, P.~Shi, K.~Li, X.~Yao, L.~Guo, and J.~Han, ``Prototype-cnn
  for few-shot object detection in remote sensing images,'' \emph{IEEE
  Transactions on Geoscience and Remote Sensing}, vol.~60, pp. 1--10, 2021.

\bibitem{xia2018dota}
G.-S. Xia, X.~Bai, J.~Ding, Z.~Zhu, S.~Belongie, J.~Luo, M.~Datcu, M.~Pelillo,
  and L.~Zhang, ``{DOTA}: A large-scale dataset for object detection in aerial
  images,'' in \emph{Proceedings of the IEEE conference on computer vision and
  pattern recognition}, 2018, pp. 3974--3983.

\bibitem{7780459}
K.~He, X.~Zhang, S.~Ren, and J.~Sun, ``Deep residual learning for image
  recognition,'' in \emph{2016 IEEE Conference on Computer Vision and Pattern
  Recognition (CVPR)}, 2016, pp. 770--778.

\bibitem{ilya2019decoupled}
I.~Loshchilov and F.~Hutter, ``Decoupled weight decay regularization,''
  \emph{arXiv preprint arXiv:1711.05101}, 2017.

\bibitem{xiong2022earthnets}
Z.~Xiong, F.~Zhang, Y.~Wang, Y.~Shi, and X.~X. Zhu, ``{EarthNets: Empowering AI
  in Earth Observation},'' \emph{arXiv preprint arXiv:2210.04936}, 2022.

\bibitem{mmdetection}
K.~Chen, J.~Wang, J.~Pang, Y.~Cao, Y.~Xiong, X.~Li, S.~Sun, W.~Feng, Z.~Liu,
  J.~Xu, Z.~Zhang, D.~Cheng, C.~Zhu, T.~Cheng, Q.~Zhao, B.~Li, X.~Lu, R.~Zhu,
  Y.~Wu, J.~Dai, J.~Wang, J.~Shi, W.~Ouyang, C.~C. Loy, and D.~Lin,
  ``{MMDetection}: Open mmlab detection toolbox and benchmark,'' \emph{arXiv
  preprint arXiv:1906.07155}, 2019.

\bibitem{zhao2021few}
Z.~Zhao, P.~Tang, L.~Zhao, and Z.~Zhang, ``Few-shot object detection of remote
  sensing images via two-stage fine-tuning,'' \emph{IEEE Geoscience and Remote
  Sensing Letters}, vol.~19, pp. 1--5, 2021.

\bibitem{wang2022context}
Y.~Wang, C.~Xu, C.~Liu, and Z.~Li, ``Context information refinement for
  few-shot object detection in remote sensing images,'' \emph{Remote Sensing},
  vol.~14, no.~14, p. 3255, 2022.

\bibitem{li2022few}
R.~Li, Y.~Zeng, J.~Wu, Y.~Wang, and X.~Zhang, ``Few-shot object detection of
  remote sensing image via calibration,'' \emph{IEEE Geoscience and Remote
  Sensing Letters}, vol.~19, pp. 1--5, 2022.

\bibitem{zhang2022few}
Y.~Zhang, B.~Zhang, and B.~Wang, ``Few-shot object detection with self-adaptive
  global similarity and two-way foreground stimulator in remote sensing
  images,'' \emph{IEEE Journal of Selected Topics in Applied Earth Observations
  and Remote Sensing}, vol.~15, pp. 7263--7276, 2022.

\bibitem{rabbi2020small}
J.~Rabbi, N.~Ray, M.~Schubert, S.~Chowdhury, and D.~Chao, ``Small-object
  detection in remote sensing images with end-to-end edge-enhanced gan and
  object detector network,'' \emph{Remote Sensing}, vol.~12, no.~9, p. 1432,
  2020.

\end{thebibliography}

\ifCLASSOPTIONcaptionsoff
  \newpage
\fi
%

\begin{IEEEbiography}[{\includegraphics[width=1in,height=1.25in,clip,keepaspectratio]{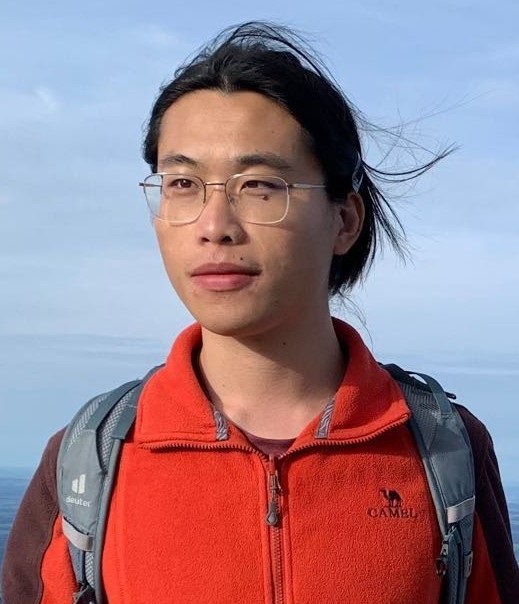}}]
{Fahong Zhang} received the B.E. degree in software engineering from Northwestern Polytechnical University, Xi'an, China, in 2017,
the M.S. degree in computer science with the Center for OPTical IMagery Analysis and Learning, Northwestern Polytechnical University, Xi'an, China.
He is now pursuing the Ph.D. degree with the department of Aerospace and Geodesy, Data Science in Earth Observation, Technical University of Munich.
His reserach interests include computer vision and satellite image processing.
\end{IEEEbiography}

\begin{IEEEbiography}[{\includegraphics[width=1in,height=1.25in,clip,keepaspectratio]{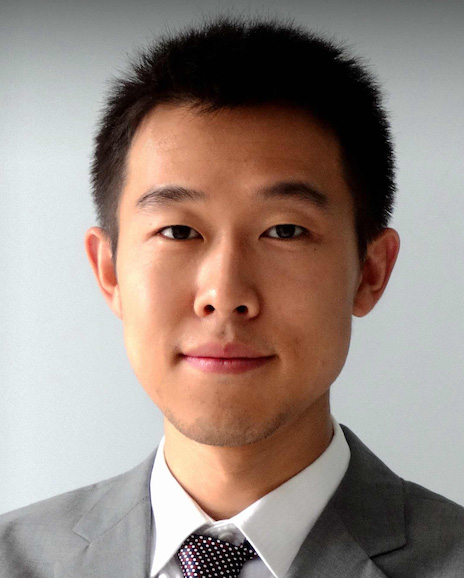}}]
{Yilei Shi} (Member, IEEE) received the Dipl.-Ing.
degree in mechanical engineering and the Dr.-Ing.
degree in signal processing from the Technical University of Munich (TUM), Munich, Germany, in
2010 and 2019, respectively.
He is a Senior Scientist with the Chair of Remote
Sensing Technology, TUM. His research interests
include fast solver and parallel computing for large-scale problems, high-performance computing and
computational intelligence, advanced methods on
synthetic aperture radar (SAR) and InSAR processing, machine learning, and deep learning for variety of data sources, such
as SAR, optical images, and medical images, and partial differential equation (PDE)-related numerical modeling and computing.
\end{IEEEbiography}

\begin{IEEEbiography}[{\includegraphics[width=1in,height=1.25in,clip,keepaspectratio]{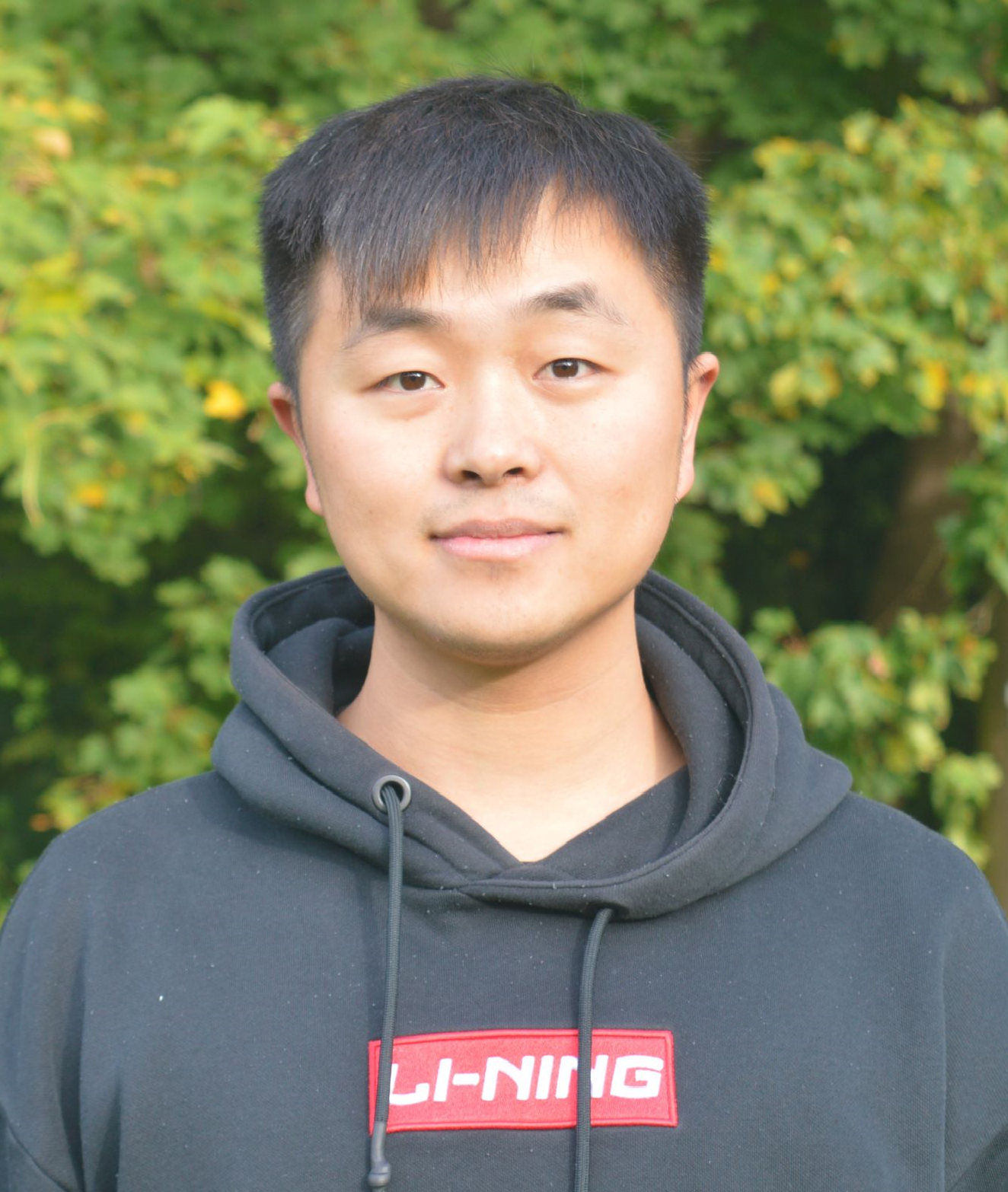}}]{Zhitong Xiong}
(Member, IEEE) received the Ph.D. degree in computer science and technology from Northwestern Polytechnical University, Xi’an, China, in 2021. He is currently a research scientist and leads the ML4Earth working group with the Data Science in Earth Observation, Technical University of Munich (TUM), Germany. His research interests include computer vision, machine learning, Earth observation, and Earth system modeling.
\end{IEEEbiography}

\begin{IEEEbiography}[{\includegraphics[width=1in,height=1.25in,clip,keepaspectratio]{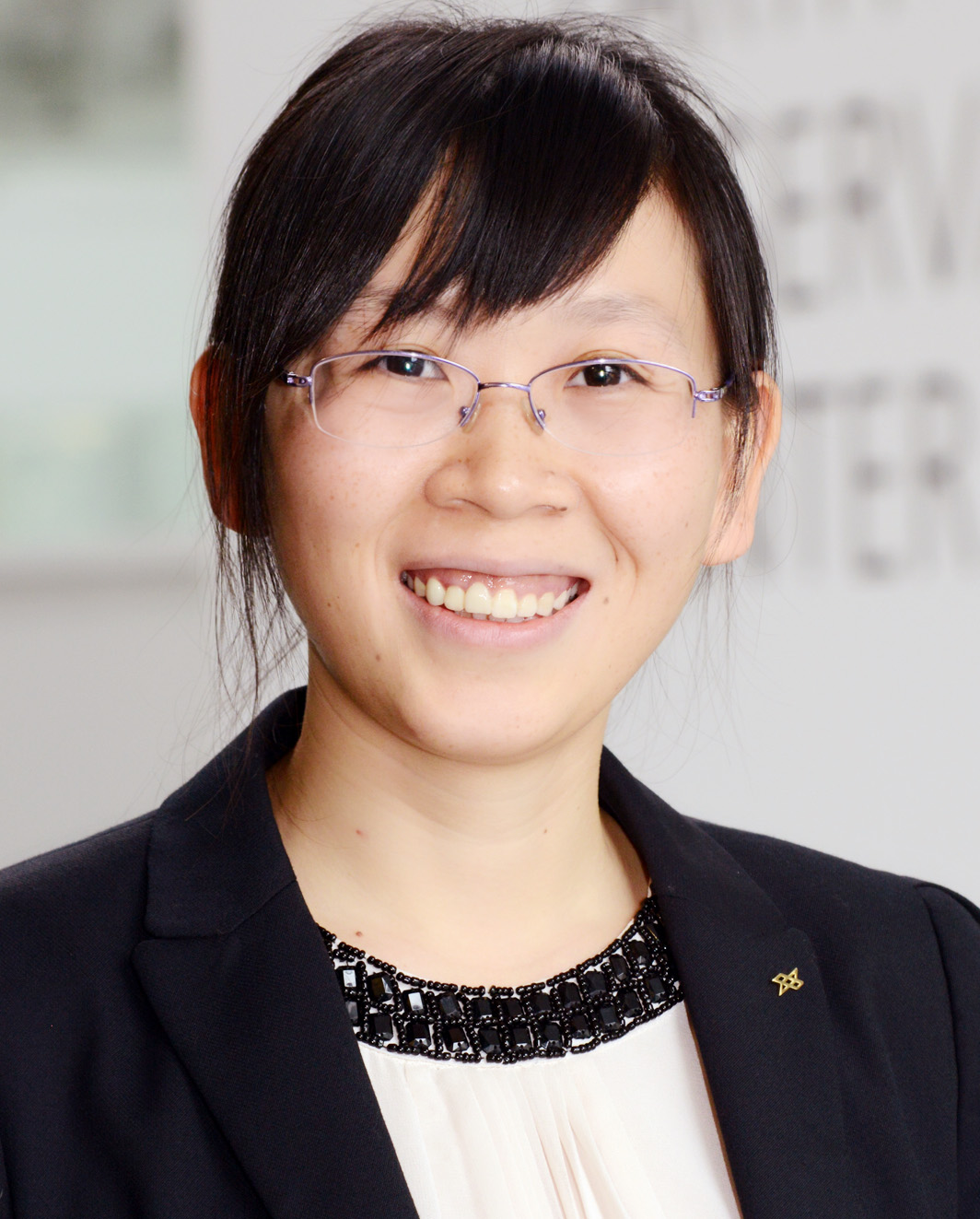}}]{Xiao Xiang Zhu}(S'10--M'12--SM'14--F'21) received the Master (M.Sc.) degree, her doctor of engineering (Dr.-Ing.) degree and her “Habilitation” in the field of signal processing from Technical University of Munich (TUM), Munich, Germany, in 2008, 2011 and 2013, respectively.
\par
She is the Chair Professor for Data Science in Earth Observation at Technical University of Munich (TUM) and was the Founding Head of the Department ``EO Data Science'' at the Remote Sensing Technology Institute, German Aerospace Center (DLR). Since 2019, Zhu is a co-coordinator of the Munich Data Science Research School (www.mu-ds.de). Since 2019 She also heads the Helmholtz Artificial Intelligence -- Research Field ``Aeronautics, Space and Transport". Since May 2020, she is the PI and director of the international future AI lab "AI4EO -- Artificial Intelligence for Earth Observation: Reasoning, Uncertainties, Ethics and Beyond", Munich, Germany. Since October 2020, she also serves as a Director of the Munich Data Science Institute (MDSI), TUM. Prof. Zhu was a guest scientist or visiting professor at the Italian National Research Council (CNR-IREA), Naples, Italy, Fudan University, Shanghai, China, the University  of Tokyo, Tokyo, Japan and University of California, Los Angeles, United States in 2009, 2014, 2015 and 2016, respectively. She is currently a visiting AI professor at ESA's Phi-lab. Her main research interests are remote sensing and Earth observation, signal processing, machine learning and data science, with their applications in tackling societal grand challenges, e.g. Global Urbanization, UN’s SDGs and Climate Change.

Dr. Zhu is a member of young academy (Junge Akademie/Junges Kolleg) at the Berlin-Brandenburg Academy of Sciences and Humanities and the German National  Academy of Sciences Leopoldina and the Bavarian Academy of Sciences and Humanities. She serves in the scientific advisory board in several research organizations, among others the German Research Center for Geosciences (GFZ) and Potsdam Institute for Climate Impact Research (PIK). She is an associate Editor of IEEE Transactions on Geoscience and Remote Sensing and serves as the area editor responsible for special issues of IEEE Signal Processing Magazine. She is a Fellow of IEEE.

\end{IEEEbiography}




\end{document}